\documentclass[journal]{IEEEtran}

\ifCLASSINFOpdf
  \usepackage[pdftex]{graphicx}
  \graphicspath{{./img/}}
  \DeclareGraphicsExtensions{.pdf,.jpeg,.png}
\else
\fi

\usepackage{amsmath}
\usepackage{array}
\usepackage{varwidth}
\usepackage{color}
\usepackage{cite}
\usepackage{amsmath}
\usepackage{graphicx}
\usepackage{amsfonts}
\usepackage{amssymb}
\usepackage{graphicx}
\usepackage{enumerate}
\usepackage{dirtytalk}
\usepackage{mathrsfs}
\usepackage{braket}
\usepackage{algorithm}
\usepackage{enumitem}
\usepackage{csquotes}
\usepackage{algpseudocode}
\usepackage{tikz}
\usetikzlibrary{arrows, arrows.meta, shapes,
automata, backgrounds, petri, matrix, calc, shapes}
\usepackage{hyperref}

\ifCLASSOPTIONcompsoc
  \usepackage[caption=false,font=normalsize,labelfont=sf,textfont=sf]{subfig}
\else
  \usepackage[caption=false,font=footnotesize]{subfig}
\fi

\usepackage{stfloats}
\ifCLASSOPTIONcaptionsoff
  \usepackage[nomarkers]{endfloat}
 \let\MYoriglatexcaption\caption
 \renewcommand{\caption}[2][\relax]{\MYoriglatexcaption[#2]{#2}}
\fi
\usepackage{url}
\DeclareMathOperator*{\argmax}{arg\,max} 

\newcommand{\todo}[1]{{\color{red}todo}}
\renewcommand{\theenumi}{\Alph{enumi}}

\newcommand{\comment}[1]{}

\tikzset{
  treenode/.style = {shape=rectangle, rounded corners,
                     draw, anchor=center,
                     text width=6.5em, align=center,
                     top color=white, bottom color=gray!30,
                     inner sep=1ex},
  decision/.style = {treenode, diamond, inner sep=0pt},
  root/.style     = {treenode
  },
  env/.style      = {treenode},
  finish/.style   = {root, bottom color=green!40},
  dummy/.style    = {}
}

\begin{document}
\bstctlcite{IEEEexample:BSTcontrol}
\title{Skill Learning by Autonomous Robotic Playing using Active Learning and Creativity}
%
%
%

\author{Simon Hangl,
		Vedran Dunjko,
		Hans J. Briegel and
		Justus Piater
\thanks{S. Hangl and J. Piater are with the Department of Computer Science,
University of Innsbruck, Austria; V. Dunjko is with the Institute of Theoretical
Physics, University of Innnsbruck, Austria and with the
Max Planck Institute for Physics Munich, Germany; H.J. Briegel is with the
Institute of Theoretical Physics, University of Innsbruck, Austria and the
Department of Philosophy, University of Konstanz, Germany.}
}

\markboth{IEEE Transactions on Robotics}%
{}
%

\maketitle
\begin{abstract}
We treat the problem of autonomous acquisition of
manipulation skills where problem-solving strategies are initially
available only for a narrow range of situations. We propose to extend
the range of solvable situations by autonomous playing with
the object. By applying previously-trained skills and behaviours,
the robot learns how to prepare situations for which a successful
strategy is already known. The information gathered during
autonomous play is additionally used to learn an environment model.
This model is exploited for active learning
and the creative generation of novel preparatory behaviours. We
apply our approach on a wide range of different
manipulation tasks, e.g.\ book grasping, grasping of
objects of different sizes by selecting different grasping strategies,
placement on shelves, and tower disassembly. We show that the
creative behaviour generation mechanism enables the robot to
solve previously-unsolvable tasks, e.g.\ tower
disassembly. We use success statistics gained during real-world
experiments to simulate the convergence behaviour of our system.
Experiments show that  active improves
the learning speed by around 9 percent in the book grasping
scenario.
\end{abstract}


\IEEEpeerreviewmaketitle

\section{Introduction}
\label{sec:intro}
\IEEEPARstart{H}{umans} perform complex object manipulations so effortlessly
that at first sight it is hard to believe that this problem is still unsolved in modern
robotics. This becomes less surprising if one considers how many
different abilities are involved in human object manipulation. These abilities
span from control (e.g. moving arms and fingers, balancing the body),
via perception (e.g. vision, haptic feedback) to planning
of complex tasks.
Most of these are not yet solved in research
by themselves, not to speak of combining them in order to design
systems that can stand up to a comparison with humans.
However, there is research on efficiently solving specific problems (or specific classes of problems)
\cite{MullingKKP20122, meeussen2010autonomous, peginhole, Hangl-2015-ICAR, Hangl-2014-ARW}.

Not only the performance of humans is
outstanding -- most manipulation
skills are learned with a high degree of autonomy. Humans are able to
use experience and apply the previously learnt lessons to
new manipulation problems.
In order to take a step towards human-like robots
we introduce a novel approach for autonomous learning that makes
it easy to embed state-of-the-art research on specific manipulation
problems. Further we aim to combine these methods in a unified framework which
autonomously learns how to combine those methods and to solve
increasingly complex tasks.

In this work we are inspired by the behaviour of infants at an age between 8 to 12 months.
Piaget identified different phases of infant
development \cite{piaget}. A phase of special interest 
is the \emph{coordination of secondary schemata} which he identifies
as the stage of \say{first actually intelligent behaviour}.
At this stage infants combine skills that were learned earlier
in order to achieve more complex tasks, e.g.
kicking an obstacle out of the way such that an object can be grasped.
Children do not predict the outcome of actions and check the corresponding
pre- and post conditions as it is done in many planning
systems \cite{5238617, Ferrein2008980, 1570410}.
To them it is only important to know
that a certain combination of manipulations is sufficient to achieve
a desired task.
The environment is prepared such that the actual skill can be
applied without a great need for generalisation.
Even adults exhibit a similar behaviour, e.g. in sports.
A golf or tennis player will always try to perform the swing
from similar positions relative to the ball.
She will position herself accordingly instead of
generalizing the swing from the current position.
This is equivalent to concatenating two behaviours,
walking towards the ball and executing the swing.

In previous work we introduced an approach that is loosely
inspired by this paradigm \cite{Hangl-2016-IROS}.
The robot holds a set of \emph{sensing actions},
\emph{preparatory behaviours} and \emph{basic behaviours}, i.e.
behaviours that solve a certain task in a narrow range of situations.
It uses the sensing actions to determine the state of the environment.
Depending on the state, a preparatory behaviour is used to bring
the environment into a state in which the task can be fulfilled
by simple replay of the basic behaviour.
The robot does not need to learn how to generalise a basic behaviour
to every possibly observable situation.
Instead, the best combination of sensing actions and preparatory behaviours
is learned by autonomous playing.

We phrase the playing as a reinforcement learning (RL) problem,
in which each rollout consists of the execution of a sensing action,
a preparatory behaviour and the desired basic behaviour.
Each rollout is time consuming, but not necessarily useful.
If the robot already knows well what to do in a specific
situation, performing another rollout in this situation does not
help to improve the policy.
However, if another situation is more interesting,
it can try to prepare it and continue the play, i.e. \emph{active learning}.
Our original approach is model-free, which makes it impossible
to exhibit such a behaviour.
In this paper we propose to learn a forward model of the
environment which allows the robot to perform transitions from
\emph{boring} situations to \emph{interesting} ones.
Another issue is the strict sequence of phases:
\emph{sensing} $\rightarrow$ \emph{preparation} $\rightarrow$ \emph{basic behaviour}.
In this work we weaken this restriction by enabling the robot
to creatively generate novel preparatory behaviours composed of
other already known behaviours.
The environment model is used to generate composite behaviours that are
potentially useful instead of randomly combining behaviours.

We illustrate the previously described concepts with the example
of book grasping.
This task is hard to generalise but easy to solve
with a simple basic behaviour in a specific situation.
The robot cannot easily get its fingers underneath the book in order to grasp it.
In a specific pose, the robot can squeeze the book between two hands,
lifting it at the spine and finally slide its fingers below the
slightly-lifted book.
Different orientations of the book would require adaption of the trajectory.
The robot would have to develop some understanding of the physical properties,
e.g. that the pressure has to be applied on the spine and that the
direction of the force vector has to point towards the supporting hand.
Learning this degree of understanding from scratch is a very hard problem.

Instead, we propose to use preparatory behaviours,
e.g. \emph{rotating the book by
$0^{\circ}$, $90^{\circ}$, $180^{\circ}$ or $270^{\circ}$},
in order to move it to the correct orientation ($\phi = 0^{\circ}$) before
the basic behaviour is executed.
The choice of the preparatory behaviour depends on the book's orientation,
e.g. $\phi \in \{ 0^{\circ}, 90^{\circ}, 180^{\circ}, 270^{\circ} \}$.
The orientation can be estimated by sliding along the book's surface, but
not by poking on top of the book.
The robot plays with the object and tries different combinations
of sensing actions and preparatory behaviours.
It receives a reward after executing the basic behaviour
and continues playing.
After training, the book grasping skill can be used as preparatory behaviour
for other skills in order to build hierarchies.

If the robot already knows well that it has to perform the behaviour
\emph{rotate $90^{\circ}$} if $\phi = 270^{\circ}$ and is confronted
with this situation again, it cannot learn anything any more,
i.e. it is \emph{bored}.
It can try to prepare a more interesting state, e.g. $\phi = 90^{\circ}$ by
executing the behaviour \emph{rotate $180^{\circ}$}.
Further, if only the behaviour \emph{rotate $90^{\circ}$} is available,
the robot cannot solve the situations $\phi \in \{90^{\circ}, 180^{\circ} \}$
by executing a single behaviour.
However, it can use behaviour compositions in order to generate
the behaviours \emph{rotate $180^{\circ}$} and \emph{rotate $270^{\circ}$}.
\section{Related Work}
\subsection{Skill chaining and hierarchical reinforcement learning}
Sutton et al. introduced the \emph{options} framework for skill learning
in a RL setting \cite{sutton1999between}.
Options are actions of arbitrary complexity, e.g. atomic actions
or high-level actions such as grasping, modelled by semi-Markov
decision processes (SMDP).
They consist of an option policy, an initiation set
indicating the states in which the policy can be executed,
and a termination condition that defines the probability of
the option terminating in a given state.
Options are orchestrated by Markov decision processes (MDP), which
can be used for planning to achieve a desired goal.
This is related to our notion of behaviours, however,
behaviours are defined in a loser way.
Behaviours do not have an initiation set and an explicit termination condition.
Behaviours are combined
by grounding them on actual executions by playing instead
of concatenating them based on planning.
Konidaris and Barto embedded so called \emph{skill chaining}
into the options framework \cite{konidaris2009skill}.
Similar to our work, options are used to bring the environment to a state
in which follow-up options can be used to achieve the task.
This is done by standand RL techniques such as Sarsa and Q-learning.
The used options themselves are autonomously generated, however,
as opposed to our method, the state space is pre-given and shared
by all options.
Instead of autonomously creating novel options, Konidaris et.
al. extended this approach by deriving options from segmenting trajectories
trained by demonstration \cite{konidaris2011robot}.
On a more abstract level, Colin et al. \cite{Colin2016196} investigated
creativity for problem-solving in artificial agents
in the context of hierarchical reinforcement learning by
emphasising parallels to psychology.
They argue that hierarchical composition of behaviours allows an agent
to handle large search spaces in order to exhibit creative behaviour.
\subsection{Model-free and model-based reinforcement learning in robotics}
Our work combines a model-free playing system and a model-based
creative behaviour generation system based on the environment model.
Work on switching between model-free and
model-based controllers was proposed in many areas of robotics
\cite{daw2005uncertainty, keramati2011speed, dolle2010path,
renaudo2014design, renaudo2015criteria, caluwaerts2012neuro,
caluwaerts2012biologically}.
The selection of different controllers is typically done by
measuring the uncertainty of the controller's predictions.
Renaudo et al. proposed switching between so called model-based
and model-free experts, where the model is learned over time.
The switching is done randomly \cite{renaudo2014design},
or by either majority vote, rank vote, Boltzmann Multiplication
or Boltzmann Addition \cite{renaudo2015criteria}.
Similar work has been done in a navigation task by Caluwaerts et al. \cite{caluwaerts2012neuro, caluwaerts2012biologically}.
Their biologically inspired approach uses three different experts,
namely a taxon expert (model-free), a planning expert (model-based),
and an exploration expert,
i.e. exploring by random actions.
A so called \emph{gating network} selects the best expert in a given situation.
All these methods hand over the complete control either to a model-based
or a model-free expert.
In contrast, our method always leaves the control with the model-free
playing system which makes the final decision on which behaviours
should be executed.
The model-based system, i.e. behaviour generation using the environment model,
is used to add more behaviours for model-free playing.
This way, the playing paradigm can still be maintained while enabling
the robot to come up with more complex ideas in case the task cannot
be solved by the model-free system alone.

Dezfouli and Balleine sequence actions and group
successful sequences to so-called \emph{habits} \cite{dezfouli2012habits}.
Roughly speaking, task solutions are generated by a dominant model-based RL system
and are transformed to atomic habits if they were rewarded many times together.
In contrast, the main driving component of our method is a model-free RL
system which is augmented with behavioural sequences by a model-based system.
This way, the robot can deal with problems without requiring an
environment model while still being able to benefit from it.
\subsection{Developmental robotics}
Our method shares properties with approaches in developmental robotics.
A common element is the concept of \emph{lifelong learning},
in which the robot develops more and more complex skills by
interacting with the environment autonomously.
W\"org\"otter et. al. proposed the concept of structural bootstrapping
\cite{structuralbootstrapping} in which knowledge acquired in earlier stages
of the robot's life is used to speed up future learning.
Weng provides an general description of a
\emph{self-aware and self-affecting agent} (SASE) \cite{weng2004developmental}.
He describes an agent with \emph{internal} and \emph{external}
sensors and actuators respectively.
It is argued that autonomous developmental robots need to be SASE agents
and concrete implementations are given, e.g. navigation or speech learning.
Our concept of boredom is an example of a pardigm,
in which the robot decides on how to procede based on \emph{internal sensing}.
In general, developmental robotics shares some key
concepts with our method, e.g. lifelong learning,
incremental development or internal sensing.
For a detailed discussion we refer to a survey by Lungarella et al. \cite{devrob-survey}.
\subsection{Active learning in robotics}
In active learning the agent can execute actions which have an impact
on the generation of training data \cite{thrun1995exploration}.
In the simplest case, the agent explores the percept-action space by random
actions \cite{whitehead2014complexity}.
The two major active learning paradigms, i.e. query-based and exploration-based
active learning, differ in the action selection mechanism.
Query-based learning systems request samples, e.g. by asking a supervisor for it.
Typically, the request is based on the agent's uncertainty
\cite{atlas1990training, cohn1994neural, cohn1996active}.
Chao et al. adopt query-based active learning for
\emph{socially guided machine learning} in robotics \cite{chao-activelearning}.
Task models are trained by interaction with a human teacher,
e.g. classifying symbols assigned to tangram compounds.
The robot could prepare a desired sample by itself,
i.e. arranging interesting tangram compounds
and asking the teacher for the class label.
In contrast to our method, this is not done in practice,
but the robot describes the desired compound.

Exploration-based active learning paradigms, on the other hand, select
actions in order to reach states with maximum uncertainty
\cite{schaal1994assessing, kaelbling1993learning,
conf/aaai/KoenigS93, sutton1990integrated}.
Salganicoff et al. \cite{salganicoff1996active} and Morales et al.
\cite{morales-grasp} used active learning for grasping.
It was used to learn a prediction model of how good
certain grasp types will work in a given situation.
All these works deal with how to select actions such that a
model of the environment can be trained more effectively.
In our approach the training of the environment model
is not the major priority.
It is a side product of the autonomous play and is used
to speed up learning and creatively generate behaviours
on top of the playing system.

Kroemer et al. \cite{kroemer2010combining} suggested a hybrid approach
of active learning and reactive control for robotic grasping.
Active learning is used to explore interesting poses using an
\emph{upper confidence bound} (UCB) \cite{sutton1998reinforcement}
policy that maximises the \emph{merit},
i.e. the sum of the expected reward mean and variance.
The actual grasps are executed by a reactive controller based on
\emph{dynamic movement primitives} (DMPs) \cite{schaal2006dynamic}
using attractor fields to move the hand towards the object
and detractor fields for obstacle avoidance.
This approach is tailored to a grasping task, in which the autonomous
identification of possible successful grasps is hard due
to high-dimensional search spaces.
In contrast, our approach is acting
on a more abstract level in which the described grasping method can be
used as one of the preparatory behaviours.
A more detailed investigation of active learning is outside the
scope of this paper and can be found in a survey by Settles \cite{settles2010active}.
Special credit shall be given to work on \emph{intrinsic motivation} 
\cite{oudeyer-iac, oudeyer1-riac, Ugur-2016-TCDS,
stoytchev2004incorporating, barto2004intrinsically, schembri2007evolution}.
It is a flavour of active learning which is commonly applied in autonomous robotics.
Instead of maximising the uncertainty, these methods try to optimise for
intermediate uncertainty.
The idea is to keep the explored situations simple enough to be able to learn
something, but complex enough to observe novel properties.
Schmidhuber provides a sophisticated summary of work on intrinsic
motivation and embedds the idea into a general framework
\cite{schidhuber-creativity-formal}.
He states that many of these works optimise some sort of
intrinsic reward, which is related to the improvement of the
prediction performance of the model.
This is closely related to our notion of boredom, in
which the robot rejects the execution of skills in a well-known situation
for the sake of using to time on improving the policy in other situations.
He further argues that such a general framework can explain concepts
like creativity and fun.
\subsection{Planning}
Many of the previously mentioned methods are concerned with training
forward models, which in consequence are used for planning in order
to achieve certain tasks.
Ugur et al. proposed a system that first learns action effects
from interaction with the objects and is trained to predict single-object
cagetories from visual perception \cite{ugur2015bottom}.
In a second stage, multi-object interaction effects are learned by
using the single-object categories, e.g. two solid objects can be
stacked on top of each other.
Discrete effects and categories are transformed into a PDDL description.
Symbolic planning is used to create complex manipulation plans,
e.g. for creating high towers by stacking.
Konidaris et al. suggest a method in which symbolic state representations
are completely determined by the agent's environment and actions \cite{konidaris2014constructing}.
They define a symbol algebra on the states derived from executed actions
that can be used for high-level planning in order to reach a desired goal.
Konidaris et al. extend this set-based formulation to a
probabilistic representation in order to deal with the uncertainty
observed in real-world settings \cite{konidaris2015symbol}.
A similar idea is present in our model-free approach, where
the selection of sensing actions
and the semantics of the estimated states depends on the desired skill.

All these approaches provide a method to build a bridge from messy
sensor data and actions to high-level planning systems for aritifial intelligence.
In order to so, similar to our approach, abstract symbols are used.
However, these systems require quite powerful machinery in order to
provide the required definition of pre- and post conditions for planning.
In our approach the robot learns a task policy directly,
which is augmented by a simple planning-based method for
creative behaviour generation.
\section{Problem Statement}
The goal is to increase the scope of situations
in which a \emph{skill} can be applied by exploiting \emph{behaviours}.
A behaviour $b \in B$ maps the complete (and partially unknown) state of system
$\mathbf{e} \in A \times E$
to another state $\mathbf{e'} \in A \times E$ with
\begin{equation}
	b : A \times E \mapsto A \times E
\label{eq:behaviour}
\end{equation}
The sets $A, \, E$ denote the internal state of the robot
and the external state of the environment (e.g. present objects) respectively.
We aim for autonomous training of a goal-directed behaviour,
i.e. a \emph{skill}.
This requires a notion of success, i.e. by a success predicate.
We define a skill $\sigma = (b^{\sigma}, \, \text{Success}^{\sigma})$ as a pair of a
\emph{basic behaviour} $b^{\sigma}$, i.e. a behaviour that solves the task
in a narrow range situations, and a predicate
\begin{equation}
	\text{Success}^{\sigma} \left( b^{\sigma} \left( \mathbf{e} \right) \right) = true
\end{equation}
with $\mathbf{e} \in D^{\sigma}$.
The non-empty set $D^{\sigma} \subseteq A \times E$ is the set of all states
in which the skill can be applied successfully, i.e. all states in
which the fixed success predicate holds.
We call the set $D^{\sigma}$ the \emph{domain of applicability} of
the skill $\sigma$.
The goal is to \emph{extend the domain of applicability} by finding behaviour compositions
$b_l \circ \dots \circ b_2 \circ b_1$ with the property
\begin{equation}
	\text{Success}^{\sigma} \left( b_l \circ \dots \circ b_2 \circ b_1 \circ b^{\sigma} \left( \mathbf{e} \right) \right) = true
\label{eq:domainextension}
\end{equation}
with $b_i \in B$ and $\mathbf{e} \in D'^{\sigma} \subseteq A \times E$ such
that $D'^{\sigma} \supsetneq D^{\sigma}$, i.e. the domain of applicability is \emph{larger} than before.
A behaviour composition $b_l \circ \dots \circ b_2 \circ b_1 \circ b^{\sigma}$
is a behaviour itself and therefore can be used to extend the domain of
applicability of other skills.
This way, skills can become more and more complex over time 
by constructing skill hierarchies.
\section{Contribution}
We extend an approach for skill learning by autonomous
playing introduced by Hangl et al. \cite{Hangl-2016-IROS}.
It uses only one preparatory behaviour per state,
i.e. allowing only behaviour compositions of length
$l = 1$, c.f. equation~\ref{eq:domainextension}.
This limitation enables the robot to perform model-free exploration due
to the reduced search space.
Allowing behaviour compositions of length $l > 1$
causes the learning problem to be intractable, but would help
to solve more complex tasks.

Approaches dealing with problems of this complexity
have to strongly reduce the search space, e.g.\
by symbolic planning
\cite{ugur2015bottom, konidaris2014constructing, konidaris2015symbol}.
We do not follow a planning-%
based paradigm in the traditional sense.
The playing-based exploration of actions remains the core component
of the system.
In order to allow behaviour compositions of length $l > 1$
while still keeping the advantage of a small search space,
we introduce a separate model-based system which generates
potentially useful behaviour compositions.
A forward model of the environment is trained with
information acquired during autonomuous play.
The environment model is used to generate new behaviour
compositions that might be worth to be tried out.
The ultimate decision whether a behaviour composition
is used, however, is still up to the playing-based system.
This way, the advantages of model-free and model-based
approaches can be combined:
\begin{enumerate}
	\item Behaviour compositions of arbitrary length
can be explored without having to deal with the combinatorial
explosion of possible behaviour compositions.
	\item No or only weak modelling of the environment is required
	because the playing-based approach alone is still stable and fully-functional.
	\item Exploration beyond the modelled percept-action space can still
	be done, e.g. a book flipping action can be used to open a box
	\cite{Hangl-2016-IROS}.
\end{enumerate}
Proposals for novel preparatory behaviours are considered proportional
to their expected usefulness.
This enables the robot to first consider more conservative plans and to
explore more unorthodox ideas in later stages.
We refer to this procedure as \emph{creative generation of behaviour proposals}.
We relate to a principal investigation of creative machines
\cite{briegel2012creative}, in which robots use a memory to
propose combinations of previous experiences in order to exhibit
\emph{new} behavioural patterns.

We further exploit the environment model for speeding up the learning process
by \emph{active learning}.
The robot can be \emph{bored} of certain situations and is not only
asking for different situations but also prepares them by itself.
Whether or not the robot is bored is part of the internal state
$\mathbf{e}_A \in A$ of the robot, which is made explicit in equation~\ref{eq:behaviour}.

We believe that a lifelong learning robot must go through different
developmental stages of increasing complexity.
Optimally, these stages are not hard-coded to the system but
emerge automatically over the course of the robot's life.
We extend our original system such that these additional mechamisms
are exploited as soon as the robot is ready for it,
i.e. the environment model is mature enough.
\section{Preliminaries}
For better understanding of the remainder of the paper, we
introduce the concept of perceptual states.
We further provide a brief description of the core reinforcement
learning method used in this paper -- projective simulation (PS) \cite{Briegel2012}.
\subsection{Perceptual states}
\label{sec:perceptualstates}
Let $\mathbf{e} \in A \times E$ be the complete physical
state of the environment.
In practice, it is impossible to estimate $\mathbf{e}$.
However, only a facet of $\mathbf{e}$ is required to
successfully perform a task.
We use haptic exploration in order to estimate the relevant
fraction of $\mathbf{e}$.
A predefined set of sensing actions $S$ is used to gather information.
For many tasks
only one sensing action $s \in S$ is required to estimate the relevant
information, e.g. the book's orientation can be determined by sliding
along the surface.
While the sensing action $s$ is executed, a multi-dimensional sensor data time series
$M = \{\mathbf{t}_{\tau}\}$ of duration $T$
with $\tau \in [1, \dots, T]$ is measured.
This time series is not the result of a deterministic process
but follows an unknown probability distribution
$p\left(M \, | \, \mathbf{e}, s\right)$.

In general, every state $\mathbf{e} \in A \times E$ potentially
requires a different action to achieve the task successfully,
e.g. how to grasp an object depends on the object pose.
However, in many manipulation problems, similar states require
a similar or even the same action.
In these cases the state space can be divided into discrete classes $e$,
e.g. the four orientations of a book in the book grasping task.
We call such a class a \emph{perceptual state}, denoted $e \in E^s_{\sigma}$.
Note that the perceptual state space $E^s_{\sigma}$ is not to be confused with
the state space of environment $E$.
The probability $p\left(e \, | \, M, s, \sigma\right)$ of a perceptual
state $e$ to be present depends on the measured sensor data $M$, the
sensing action $s$ and the skill $\sigma$ for which the sensing
action $s$ is used, e.g. poking in book grasping means
something different than in box opening.
The perceptual state spaces of two sensing actions
$s, s' \in S$ can coincide, partly overlap or be distinct
e.g. sliding along the surface allows the robot
to estimate the orientation of a book, whereas poking does not.
\subsection{Projective simulation}
\label{sec:projectivesimulation}
Projective simulation (PS) \cite{Briegel2012} is a
framework for the design of intelligent agents and can be
used for reinforcement learning (RL).
PS was shown to exhibit competitive performance in several reinforcement learning scenarions ranging
from classical RL problems to adaptive quantum computation
\cite{melnikov2014projective, mautner2015projective, melnikov2015projective, tiersch2015adaptive}.
It is a core component of our method and was chosen due to structural advantages,
conceptual simplicity and good extensibility.
We brief\-ly describe the basic concepts and the modifications
applied in this paper.
A detailed investigation of its properties can be
found in \cite{mautner2015projective}.
\begin{figure}[t!]
\centering
  \scalebox{0.67} {
\begin{tikzpicture}[every node/.style = {shape=rectangle, rounded corners},>=stealth',bend angle=45, auto]

  \tikzstyle{place}=[circle,thick,draw=blue!75,fill=blue!20,minimum size=6mm]

  \begin{scope}
                
    \draw(0,1) node[ellipse, minimum height=1.3cm,draw] (c1) {Clip 1 $\equiv c_{p_1}$ (Percept)};
    \draw(0,-1.5) node[ellipse, minimum height=1.3cm,draw] (c2) {Clip 2 $\equiv c_{p_2}$ (Percept)};
    
    \draw(4,2) node[ellipse, minimum height=1cm,draw] (c3) {Clip 3};   
    \draw(3.8,0.2) node[ellipse, minimum height=1cm,draw] (c4) {Clip 4};
    \draw(4,-2) node[ellipse, minimum height=1cm,draw] (c5) {Clip 5};
    
    \draw(6.5,0) node[ellipse, minimum height=1cm,draw] (c6) {Clip 6};
    
    \draw(8,2) node[ellipse, minimum height=1.3cm,draw] (c7) {Clip 7 $\equiv b_1$ (Behaviour)};
    \draw(8,-2) node[ellipse, minimum height=1.3cm,draw] (c8) {Clip 8 $\equiv b_2$ (Behaviour)};

	\path[->] (c1) edge [line width=0.4mm] node {$p_{1 \rightarrow 3}$} (c3);
	\path[->] (c1) edge [line width=0.4mm] node {$p_{1 \rightarrow 4}$} (c4);
	\path[->] (c1) edge [line width=0.4mm] node {$p_{1 \rightarrow 5}$} (c5);
	
	\path[->] (c2.north) edge [line width=0.4mm] node {$p_{2 \rightarrow 4}$} (c4);
	\path[->] (c2) edge [line width=0.4mm] node [below] {$p_{2 \rightarrow 5}$} (c5);
	\path[->] (c6) edge [line width=0.4mm] node {$p_{6 \rightarrow 3}$} (c3);
	\path[->] (c5) edge [line width=0.4mm] node {$p_{5 \rightarrow 6}$} (c6);
	
	\path[->] (c4) edge [line width=0.4mm] node [below] {$p_{4 \rightarrow 6}$} (c6);
	
	\path[->] (c5) edge [line width=0.4mm] node {$p_{5 \rightarrow 8}$} (c8);
	\path[->] (c6) edge [line width=0.4mm] node {$p_{6 \rightarrow 8}$} (c8);
	\path[->] (c3) edge [line width=0.4mm] node {$p_{3 \rightarrow 7}$} (c7);
	\path[->] (c3) edge [line width=0.4mm] node [left] {$p_{3 \rightarrow 4}$} (c4);
	\path[->] (c6) edge [line width=0.4mm] node {$p_{6 \rightarrow 7}$} (c7);
	
  \end{scope}
\end{tikzpicture}
}
\caption{Exemplary sketch of an episodic and compositional memory (ECM).
A random walk always starts at a percept clip (e.g. clip 1, clip 2)
and ends with an behaviour clip (e.g. clip 7, clip 8).
A transition $c \rightarrow c'$ from clip $c$ to clip $c'$ is
done with the probability $p_{c \rightarrow c'}$.}
\label{fig:ecm}
\end{figure}
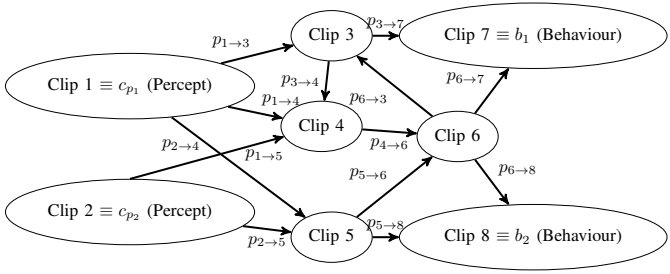

Roughly speaking, the PS agent learns the probability
distribution $p\left(b \, | \, \mathbf{\lambda}, \, \mathbf{e} \right)$
of executing a behaviour $b$ (e.g. a preparatory behaviour)
given the observed sensor data $\mathbf{\lambda}$
(e.g. a verbal command regarding which skill to execute) in order to
maximise a given reward function $r\left( b, \mathbf{\lambda}, \, \mathbf{e} \right)$.
In this paper, reward is given if
$\text{Success}^{\sigma}( b \circ b^{\sigma} \left( \mathbf{e} \right) ) = \text{true}$,
given a command $\lambda$ to execute skill $\sigma$ in
the present environment state $\mathbf{e}$.
Note that the state $\mathbf{e}$ is never observed directly.
Instead, perceptual states are estimated throughout the skill execution.

In general, the core of the PS agent is the so-called
\emph{episodic and compositional memory} (ECM).
An exemplary sketch of an ECM is shown in Fig. \ref{fig:ecm}.
It stores fragments of experience, so-called \emph{clips}, and connections between them.
Each clip represents a previous experience, i.e. percepts and actions.

The distribution $p\left(b \, | \, \mathbf{\lambda}, \, \mathbf{e} \right)$
is updated after a rollout, i.e. observing a percept, choosing and
executing a behaviour according to
$p\left(b \, | \, \mathbf{\lambda}, \, \mathbf{e} \right)$,
and receiving reward from the environment.
The distribution $p\left(b \, | \, \mathbf{\lambda}, \, \mathbf{e} \right)$
is implicitly specified by assigning transition probabilities
$p_{c \rightarrow c'} = p \left( c' \, | \, c \right)$
to all pairs of clips
$\left( c, \, c' \right)$ (in Fig. \ref{fig:ecm}
only transitions with probability
$p_{c \rightarrow c'} \neq 0$ are visualised).
Given a certain \emph{percept clip},
i.e. a clip without inbound transitions like clips 1 and 2,
the executed \emph{behaviour clip},
i.e. a clip without outbound transitions like clips 7 and 8,
is selected by a \emph{random walk} through the ECM.
A random walk is done by hopping from clip to clip according to the
respective transition probabilities until a behaviour is reached.
Clips are discrete whereas sensor data is typically continuous, e.g. voice commands.
A domain-specific input coupler distribution $I(c_p \, | \, \mathbf{\lambda}, \mathbf{e})$ 
modelling the probability of observing a discrete percept clip $c_p$
given an observed signal $\mathbf{\lambda}$ is required.
The distribution $p\left(b \, | \, \mathbf{\lambda}, \mathbf{e} \right)$
is given by a random walk through the ECM with
\begin{equation}
p\left(b \, | \, \mathbf{\lambda}, \mathbf{e} \right) = \sum_{c_p}{ \left( I(c_p \, |
\, \mathbf{\lambda}, \mathbf{e}) \sum_{w \in \Lambda(a, c_p)}{ p(b \, | \, c_p, w) } \right) }
\end{equation}
where $p(b \, | \, c_p, w)$ is the probability
of reaching behaviour $b$ from percept $c_p$ via the path
$w = \left( c_p = c_{1}, c_{2}, \dots, c_{K} = b \right)$.
The set $\Lambda(b, c_p)$ is the set of all paths from the percept clip $c_p$ to
the behaviour clip $b$.
The path probability is given by
\begin{equation}
p(b \, | \, c_p, w) = \prod_{j = 1}^{K - 1} p{\left( c_{j + 1} \text{ } | \text{ } c_{j} \right)}
\end{equation}

The agent learns by adapting the probabilities $p_{c \rightarrow c'}$
according to the received reward (or punishment)
$r \in \mathbb{R}$.
The transition probability $p_{c \rightarrow c'}$
from a clip $c$ to another clip $c'$ is specified by
the abstract transition weights $h \in \mathbb{R}^+$ with
\begin{equation}
p_{c \rightarrow c'} = p \left( c \text{ } | \text{ } c' \right) = \frac{h_{c \rightarrow c'}}{\sum_{\hat{c}} h_{c \rightarrow \hat{c}}}
\label{equ:psrandomwalk}
\end{equation}
After each rollout, all weights $h_{c \rightarrow c'}$
are updated.
Let $w$ be a random walk path with reward $r^{(t)} \in \mathbb{R}$
at time $t$.
The transition weights are updated according to
\begin{equation}
h^{t + 1}_{c \rightarrow c'}
= \text{max}\left(1, \, h^t_{c \rightarrow c'}
 - \zeta \left(h^t_{c \rightarrow c'} - 1 \right)
 + \rho \left( c, c', w \right) r^{(t)}\right)
\label{equ:pslearn}
\end{equation}
where $\rho(c, c', w)$ is 1 if the path $w$ contains the
transition $c \rightarrow c'$ and 0 otherwise.
The \emph{forgetting factor} $\zeta$
defines the rate with which the agent forgets previously
learned policies.
\section{Skill learning by robotic playing}
\label{sec:origapproach}
The following section describes the method for autonomous skill
acquisition by autonomous playing on which this work is based on
\cite{Hangl-2016-IROS}.
The sections \ref{sec:envmodel} -- \ref{sec:creativity}
present extensions that run in parallel
and augment the autonomous playing.
\begin{figure*}[t!]
\centering
  \scalebox{0.625} {
\begin{tikzpicture}[every node/.style = {shape=rectangle, rounded corners},>=stealth',bend angle=45, auto, font=\fontsize{11}{20}\selectfont]

  \tikzstyle{place}=[circle,thick,draw=blue!75,fill=blue!20,minimum size=6mm]

  \begin{scope}

	\node[text width=5cm,align=left] at (-2.3,7.0) {Layer A $\equiv$ \\ Input couplers};
	
	\node[text width=5cm,align=left] at (-2.3,3.7) {Layer B $\equiv$ \\ Desired skills};
	
	\node[text width=5cm,align=left] at (-2.3,1.5) {Layer C $\equiv$ \\ Sensing actions};
	
	\node[text width=5cm,align=left] at (-2.3,-1.0) {Layer D $\equiv$ \\ Perceptual \\ states};
	
	\node[text width=5cm,align=left] at (-2.3,-4.2) {Layer E $\equiv$ \\ Preparatory\\ behaviours};
  
  	\draw(7,7.0) node[rectangle, minimum width=3cm, minimum height=1.5cm,draw] (i1) {$I_{\mathrm{sp}} \equiv$ Speech recognition};
  	\draw(15,7.0) node[rectangle, minimum width=3cm, minimum height=1.5cm,draw] (i2) {$I_{\mathrm{kb}} \equiv$ Keyboard input};
  
	\draw(5.7,0.9) node[rectangle, minimum width=16cm, minimum height=6.8cm,draw] (skillbox1) {};
  	\node[text width=6cm,align=left] at (1.05,3.7) {Skill $\sigma_1$};  
	\draw(5.7,3.5) node[ellipse, minimum height=1.1cm,draw] (c1) {$\sigma_1 \equiv $ Skill 1};
  
  	\draw(2.25,0.1) node[rectangle, minimum width=8cm, minimum height=4.3cm,draw]
  	(sensingbox1) {};
  	\node[text width=6cm,align=left] at (1.5,1.8) {Sensing action $s_1$};
  	\draw(2.25,1.5) node[circle, minimum size=1cm,draw] (s1) {$s_1$};
  
	\draw node[rectangle, minimum width=7cm, minimum height=2cm,draw, below of = s1,
	yshift = -1cm] (statebox1) {};
	\node[text width=6cm,align=left, below of = s1, yshift=-0.35cm, xshift=-0.2cm]
				{Perceptual states of sensing action $s_1$};
    \draw node[circle, minimum size=1cm,draw, below of = s1, yshift=-1.2cm, xshift = -2.25cm] (e11) {$e_1^{s_1}$};
    \draw node[circle, minimum size=1cm,draw, below of = s1, yshift=-1.2cm, xshift = -0.75cm] (e12) {$e_2^{s_1}$};
    \draw node[circle, minimum size=1cm,draw, below of = s1, yshift=-1.2cm, xshift = 0.75cm] (e1dot) {$e_{\dots}^{s_1}$};
    \draw node[circle, minimum size=1cm,draw, below of = s1, yshift=-1.2cm, xshift = 2.25cm] (e1L1) {$e_{L_{\sigma,s_1}}^{s_1}$};
    
    \draw node[rectangle, minimum width=3cm, minimum height=4.3cm,draw, right of = sensingbox1, xshift = 5cm] (sensingbox2) {%
    Sensing action $s_{\dots}$};
    
    \draw(11.75,0.1) node[rectangle, minimum width=3cm, minimum height=4.3cm,draw] (sensingbox3) {%
    Sensing action $s_N$};
    
    \draw(16.5,0.9) node[rectangle, minimum width=4cm, minimum height=6.8cm,draw,
    right of = skillbox1, xshift = 9.5cm] (skillbox2) {%
    Skill $\sigma_{\dots}$};
    
    \draw(21.3,0.9) node[rectangle, minimum width=4cm, minimum height=6.8cm,draw,
    right of = skillbox2, xshift = 3.5cm] (skillbox3) {%
    Skill $\sigma_{K}$};
    
    \draw(0,-4.5) node[rectangle, minimum width=3cm, minimum height=1.5cm,draw] (a1) {%
    \begin{varwidth}{3cm}
    $b_1 \equiv$ Preparatory behaviour 1
    \end{varwidth}
    };
    \draw(3.5,-4.5) node[rectangle, minimum width=3cm, minimum height=1.5cm,draw] (a2) {$\dots$};
    \draw(7.5,-4.5) node[rectangle, minimum width=4cm, minimum height=1.5cm,draw] (a3) {%
    \begin{varwidth}{4cm}
    $b_{J - 1} \equiv$ Preparatory behaviour $J - 1$
    \end{varwidth}
    };
    \draw(11.5,-4.5) node[rectangle, minimum width=3cm, minimum height=1.5cm,draw] (a4) {%
    \begin{varwidth}{3cm}
    $b_{\emptyset} \equiv$ \emph{void} behaviour
    \end{varwidth}
    };
    
    \draw(15,-4.5) node[rectangle, minimum width=3cm, minimum height=1.5cm, dashed, draw] (atocp1){};
    \node[dummy] (dummy1a) [below of = skillbox3, yshift = -4.4cm, xshift = 3cm] {};
    \node[dummy] (dummy2a) [above of = skillbox3, yshift = 3.0cm, xshift = 3cm] {};
    \node[dummy] (dummy3a) [above of = skillbox3, yshift = 3.0cm] {};
    \node[dummy] (dummy4a) [above of = skillbox2, yshift = 3.0cm] {};
    \node[dummy] (dummy5a) [above of = skillbox1, yshift = 3.0cm] {};
    \path[-] (atocp1.east) [dashed] edge  node {} (dummy1a.west);
    \path[-] (dummy1a.north) [dashed] edge  node {} (dummy2a.south);
    \path[-] (dummy2a.west) [dashed] edge  node {} (dummy3a.east);
    \path[-] (dummy3a.west) [dashed] edge  node {} (dummy4a.east);
    \path[-] (dummy4a.west) [dashed] edge  node {} (dummy5a.east);
    
    \path[->] (dummy3a.south) [dashed] edge  node {} (skillbox3.north);
    \path[->] (dummy4a.south) [dashed] edge  node {} (skillbox2.north);
    \path[->] (dummy5a.south) [dashed] edge  node {} (c1.north);
    
    \path[->] (c1.south) edge [line width=0.2mm] node {} (s1.north);
    \path[->] (c1.south) edge [line width=0.2mm] node {} (sensingbox2.north);
    \path[->] (c1.south) edge [line width=0.2mm] node {} (sensingbox3.north);
    
    \path[->] (s1.south) edge [line width=0.2mm] node {} (e11.north);
    \path[->] (s1.south) edge [line width=0.2mm] node {} (e12.north);
    \path[->] (s1.south) edge [line width=0.2mm] node {} (e1dot.north);
    \path[->] (s1.south) edge [line width=0.2mm] node {} (e1L1.north);
    
    \path[->] (e11.south) edge [line width=0.2mm] node {} (a1.north);
    \path[->] (e11.south) edge [line width=0.2mm] node {} (a2.north);
    \path[->] (e11.south) edge [line width=0.2mm] node {} (a3.north);
    \path[->] (e11.south) edge [line width=0.2mm] node {} (a4.north);
    \path[->] (e12.south) edge [line width=0.2mm] node {} (a1.north);
    \path[->] (e12.south) edge [line width=0.2mm] node {} (a2.north);
    \path[->] (e12.south) edge [line width=0.2mm] node {} (a3.north);
    \path[->] (e12.south) edge [line width=0.2mm] node {} (a4.north);
    \path[->] (e1dot.south) edge [line width=0.2mm] node {} (a1.north);
    \path[->] (e1dot.south) edge [line width=0.2mm] node {} (a2.north);
    \path[->] (e1dot.south) edge [line width=0.2mm] node {} (a3.north);
    \path[->] (e1dot.south) edge [line width=0.2mm] node {} (a4.north);
    \path[->] (e1L1.south) edge [line width=0.2mm] node {} (a1.north);
    \path[->] (e1L1.south) edge [line width=0.2mm] node {} (a2.north);
    \path[->] (e1L1.south) edge [line width=0.2mm] node {} (a3.north);
    \path[->] (e1L1.south) edge [line width=0.2mm] node {} (a4.north);
    
    \path[->] (sensingbox2.south) edge [line width=0.2mm] node {} (a1.north);
    \path[->] (sensingbox2.south) edge [line width=0.2mm] node {} (a2.north);
    \path[->] (sensingbox2.south) edge [line width=0.2mm] node {} (a3.north);
    \path[->] (sensingbox2.south) edge [line width=0.2mm] node {} (a4.north);
    \path[->] (sensingbox3.south) edge [line width=0.2mm] node {} (a1.north);
    \path[->] (sensingbox3.south) edge [line width=0.2mm] node {} (a2.north);
    \path[->] (sensingbox3.south) edge [line width=0.2mm] node {} (a3.north);
    \path[->] (sensingbox3.south) edge [line width=0.2mm] node {} (a4.north);
    \path[->] (skillbox2.south) edge [line width=0.2mm] node {} (a1.north);
    \path[->] (skillbox2.south) edge [line width=0.2mm] node {} (a2.north);
    \path[->] (skillbox2.south) edge [line width=0.2mm] node {} (a3.north);
    \path[->] (skillbox2.south) edge [line width=0.2mm] node {} (a4.north);
    \path[->] (skillbox3.south) edge [line width=0.2mm] node {} (a1.north);
    \path[->] (skillbox3.south) edge [line width=0.2mm] node {} (a2.north);
    \path[->] (skillbox3.south) edge [line width=0.2mm] node {} (a3.north);
    \path[->] (skillbox3.south) edge [line width=0.2mm] node {} (a4.north);
    
    \path[->] (i1.south) edge [line width=0.2mm] node {} (c1.north);
    \path[->] (i1.south) edge [line width=0.2mm] node {} (skillbox2.north);
    \path[->] (i1.south) edge [line width=0.2mm] node {} (skillbox3.north);
    
    \path[->] (i2.south) edge [line width=0.2mm] node {} (c1.north);
    \path[->] (i2.south) edge [line width=0.2mm] node {} (skillbox2.north);
    \path[->] (i2.south) edge [line width=0.2mm] node {} (skillbox3.north);
    
    \path[->] (e11.south) edge [line width=0.2mm,dashed] node {} (atocp1.north);
    \path[->] (e12.south) edge [line width=0.2mm,dashed] node {} (atocp1.north);
    \path[->] (e1dot.south) edge [line width=0.2mm,dashed] node {} (atocp1.north);
    \path[->] (e1L1.south) edge [line width=0.2mm,dashed] node {} (atocp1.north);
    \path[->] (sensingbox2.south) edge [line width=0.2mm,dashed] node {} (atocp1.north);
    \path[->] (sensingbox3.south) edge [line width=0.2mm,dashed] node {} (atocp1.north);
    \path[->] (skillbox2.south) edge [line width=0.2mm,dashed] node {} (atocp1.north);
    \path[->] (skillbox3.south) edge [line width=0.2mm,dashed] node {} (atocp1.north);
    
  \end{scope}
\end{tikzpicture}
}
\caption{ECM for autonomous robotic playing. For execution a random walk is
performed from layer A to layer E.
The transition from layer C to layer D is performed by executing the corresponding
sensing action $s$, measuring the haptic data and using a time series classifier.
All other transitions follow equation \ref{equ:psrandomwalk}.
The preparatory behaviour $b_{\emptyset} \equiv$ (\emph{void} behaviour) is always in the
set of preparatory behaviours.
The dashed box and lines refer to skills used as preparatory behaviours in
order to build skill hierarchies.
After preparation, the \emph{basic} behaviour $b^{\sigma}$ corresponding to the
desired skill $\sigma$ is executed.
}
\label{fig:ecmplaying}
\end{figure*}
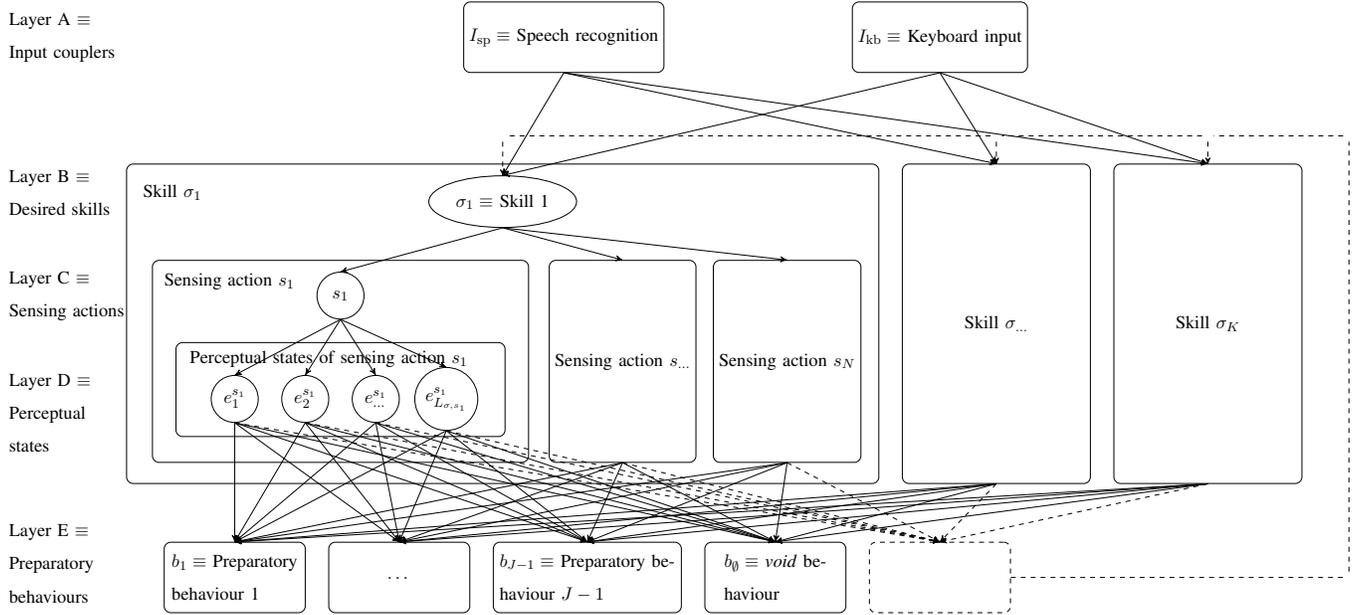
\subsection{ECM for robotic playing}
\label{sec:ecmplayingform}
A skill $\sigma$ is executed by a random walk through the layered ECM
shown in Fig. \ref{fig:ecmplaying}.
It consists of the following layers:
\begin{enumerate}
	\item Input couplers: Input couplers map user commands about which
	skill to execute to the corresponding skill clip.
	The percept of this ECM is not the state of the environment, but the command
	of which skill to execute.
	\item Desired skills: Each clip $\sigma$, i.e. a percept clip,
	represents a skill the robot is able to perform.
	\item Sensing actions: Each clip $s \in S$ corresponds to
	one sensing action.
	All skills share the same sensing actions.
	\item Perceptual states: Each clip $e \in E_{\sigma}^s$ corresponds
	to a perceptual state under the
	sensing action $s$ for the skill $\sigma$.
	Note that the perceptual states are
	different for each skill-sensing action pair $(\sigma, s)$ and
	typically do not have the same semantics, e.g. the states
	under sensing action $s \in S$ might identify the object pose, whereas
	the states under $s' \in S$ might denote the object's concavity.
	\item Preparatory behaviours: Each clip corresponds to a behaviour which can be
	atomic (solid transitions) or other trained skills (dashed
	transitions).
	Since the basic behaviour $b^{\sigma}$ of a skill was shown to the robot
	in one perceptual state,
	there is at least one state that does not require preparation.	
	Therefore, the \emph{void}-behaviour $b_{\emptyset}$, in which no preparation is done,
	is in the set of behaviours.
\end{enumerate}
The robot holds the sets of skills $\{ \sigma = \left(b^{\sigma}, \text{Success}^\sigma \right) \}$, sensing actions $S$ (e.g. sliding, poking, pressing)
and preparatory behaviours $B$ (e.g. pushing).
A skill is executed by performing a random walk through the ECM
and by performing the actions along the path.
The idle robot waits for a skill execution command $\lambda$
whic is mapped to skill clips in Layer B by
coupler functions, e.g. $I_{\mathrm{kb}}$ and
$I_{\mathrm{sp}}$ mapping a keyboard input / voice commands
to the desired skill clip $\sigma$.
A sensing action $s \in S$ is chosen and executed according
to the transition probabilities and a sensor data time series $M$ is measured.
The perceptual state $e \in E_{\sigma}^s$ is estimated from $M$.
This transition is done deterministically by a classifier
and not random as in the steps before.
Given the perceptual state $e$, the environment is prepared by executing a
behaviour $b \in B$.
Finally, the basic behaviour $b^{\sigma}$ is executed.
If a basal bevahiour of a skill requires an object
to be grasped, only the sensing action  \emph{weighing} is available
in order to estimate whether an object is grasped.
We stress that this is only a restriction enforced
due to practical considerations and is not required
in principle.
\subsection{Skill Training}
A novel skill $\sigma = (b^{\sigma}, \text{Success}^{\sigma})$ is trained
by providing the basic behaviour $b^{\sigma}$ for a narrow range of situations,
e.g. by hard coding or learning from demonstration
\cite{Hangl-2015-ICAR, Argall2009469, 614389,
asfourimitationlearning, 4399517, kormushev2011imitation, konidaris2011robot}.
The domain of applicability is extended by learning:
\begin{enumerate}[label=\alph*)]
	\item \label{enum:prob1}which sensing action should be used to estimate
	the relevant perceptual state;
	\item \label{enum:prob2}how to estimate the perceptual state from haptic data;
	\item \label{enum:prob3}which preparatory behaviour helps to achieve the task in a
	given perceptual state.
\end{enumerate}
The skill ECM (Fig. \ref{fig:ecmplaying}) is initialised in a meaningful way
(sections \ref{sec:hapticdbcreation}, \ref{sec:ecminitialisation})
and afterwards refined by executing the skills and collecting rewards, i.e.
\emph{autonomous playing}.
\subsubsection{Haptic database creation}
\label{sec:hapticdbcreation}
In a first step, the robot creates a haptic database by
exploring how different perceptual states \enquote{feel}, c.f.
problem~\ref{enum:prob2}.
It performs all sensing actions $s \in S$ several times
in all perceptual states $e^s$, acquires the sensor data
$M$ and stores the sets $\{ \left( e^s, s, \{ M \} \right) \}$.
With this database the distribution $p(e \, | \, M, s, \sigma)$
(section~\ref{sec:perceptualstates}) can be approximated and a perceptual state
classifier is trained.

There are two ways of preparing different perceptual states.
Either the supervisor prepares the different states (e.g. all four book poses)
or the robot is provided with information on how to prepare them
autonomously (e.g. \emph{rotate by 90$^{\circ}$} produces all poses).
In the latter case the robot assumes that after execution of the
behaviour a new perceptual state $e'$ is present and
adds it to the haptic database.
This illustrates three important assumptions: The state
$e^s \in E^s_{\sigma}$ is invariant under the sensing action $s \in S$ (e.g.
the book's orientation remains the same irrespective of how often
\emph{sliding} is executed)
but not under preparatory behaviours $b \in B$
(e.g. the book's orientation changes by using the \emph{rotate $90^{\circ}$}
behaviour), which yields
\begin{equation}
e^{s} \xrightarrow{s} e^{s}
\label{equ:assumptionsense}
\end{equation}
\begin{equation}
e^{s} \xrightarrow{b} e'^{s}
\label{equ:assumptionprep}
\end{equation}
Further we do not assume that a sensing action $s'$ leaves the
perceptual state $e^{s}$ of another sensing action $s$ unchanged
(e.g. sliding softly along a tower made of cups does not change
the position of the cups whereas poking from the side may cause
the tower to collapse).
This insight is reflected by the example
\begin{equation}
e^{s} \xrightarrow{s} e^{s} \xrightarrow{s} \dots \xrightarrow{s} e^{s} \xrightarrow{s'} e^{s'} \xrightarrow{s} e'^{s}
\label{equ:assumptionsense2}
\end{equation}
\subsubsection{ECM Initialisation}
\label{sec:ecminitialisation}
The ECM in Fig.~\ref{fig:ecmplaying} is initialised with the uniform
transition weights $h_{\mathrm{init}}$ except for the
weights between layers B and C.
These weights are initialised such that the agent
prefers sensing actions $s \in S$ that
can discriminate well between their environment
states $e^s \in E^s_{\sigma}$.
After the generation of the haptic
database the robot performs cross-validation for
the perceptual state classifier of each sensing
action $s \in S$ and computes the average success
rate $r_s$.
A \emph{discrimination score} $D_s$ is computed by
\begin{equation}
D_s = \exp{\left( \alpha r_{s} \right)}
\label{equ:discscore}
\end{equation}
with the free parameter $\alpha$ called \emph{stretching factor}.
The higher the discrimination score, the better the sensing
action can classify the corresponding perceptual states.
Therefore, sensing actions with a high discrimination score should be preferred
over sensing actions with a lower score.
The transition weights between all pairs of
the skill clip $\sigma$ and the sensing
action clips $s \in S$ are initialised with
$h_{\sigma \rightarrow s} = D_s$.
We use a C-SVM classifier implemented in LibSVM \cite{libsvm}
for state estimation.
\subsubsection{Extending the domain of applicability}
\label{sec:skillgeneralisation}
\begin{figure}[t!]
\centering
\scalebox{0.56} {
\begin{tikzpicture}

	\draw(11.7cm, -8.5cm) node[rectangle, minimum width=4.1cm, minimum height=11cm,draw=none,
	top color=gray!15, bottom color=gray!15]
  	(shadow1) {};
  	
  	\draw(0cm, -16cm) node[rectangle, minimum width=3.5cm, minimum height=4cm,draw=none,
	top color=gray!15, bottom color=gray!15]
  	(shadow2) {};
  	
  	\draw(4cm, -10.75cm) node[rectangle, minimum width=4cm, minimum height=5cm,draw=none,
	top color=gray!15, bottom color=gray!15]
  	(shadow3) {};

	\node[root] (init1) [minimum height=1.5cm] {Provide new basic behaviour $b^{\sigma}$};
	\node[env] (init2) [right of = init1, xshift = 3cm, minimum height=1.5cm] {Create haptic database};
	\node[env] (init3) [right of = init2, xshift = 3cm, minimum height=1.5cm] {Initialise ECM};
	
	\node[env] (exec1) [below of = init3, yshift = -3cm, minimum height=1.5cm] {Execute sensing action
	$s \in S$ and measure sensor data $M$};
	
	\node[env] (exec2) [below of = exec1, yshift = -3cm, minimum height=1.5cm] {Estimate perceptual
	state $e^s \in E^s_{\sigma}$};
	
	\node[decision] (boredomactive) [below of = exec2, yshift = -3cm] {Boredom activated?};
	\node[decision] (creativeactive) [below of = boredomactive, yshift = -3cm] {Creativity activated?};
	
	\node[env] (env1) [left of = boredomactive, xshift = -3cm] {Re-execute sensing action $s$};
	
	\node[env] (exec3) [left of = env1, xshift = -3cm, minimum height=1.5cm] {Execute
	behaviour $b \in B$};
	\node[env] (exec4) [above of = exec3, yshift = 1.5cm, minimum height=1.5cm] {Collect reward
	and update ECM};
	
	\node[env] (env2) [right of = exec4, xshift = 3cm] {Update environment model of $s$};
	
	\node[env] (creative1) [below of = exec3, yshift = -3cm] {Generate and add novel behaviours};
	\node[dummy] (d3) [below of = creativeactive, yshift = -0.5cm] {};
	\node[dummy] (d4) [below of = creative1, yshift = -0.5cm] {};
	
	\node[decision] (welltrained) [above of = exec4, yshift = 2.25cm] {Skill well-trained?};
	\node[env] (finished1) [above of = welltrained, yshift = 2.25cm, minimum height=1.5cm] {Add $\sigma$ as preparatory behaviour for other skills};
	\node[dummy] (d2) [left of = exec1, xshift = -3cm] {};
	
	\node[decision] (isbored) [right of = boredomactive, xshift = 3cm] {Is bored in
	$e^s \in E^s_{\sigma}$?};
	\node[env] (boredom1) [right of = exec2, xshift = 3cm, minimum height=1.5cm] {Perform transition to an interesting state};
	\node[dummy] (d1) [right of = exec1, xshift = 3cm] {};
	
	\node[dummy] (dbored) [right of = creativeactive, xshift = 3cm] {};
	
	\path[-{>[scale=2.5,length=2,width=4]},line width=0.4pt] (init1) edge node {} (init2);
	\path[-{>[scale=2.5,length=2,width=4]},line width=0.4pt] (init2) edge node {} (init3);
	\path[-{>[scale=2.5,length=2,width=4]},line width=0.4pt] (init3) edge node {} (exec1);
	\path[-{>[scale=2.5,length=2,width=4]},line width=0.4pt] (exec1) edge node {} (exec2);
	\path[-{>[scale=2.5,length=2,width=4]},line width=0.4pt] (exec2) edge node {} (boredomactive);
	\path[-{>[scale=2.5,length=2,width=4]},line width=0.4pt] (boredomactive) [dashed] edge node [above] {yes} (isbored);
	\path[-,line width=0.4pt] (isbored.south) [dashed] edge node [left] {no} (dbored);
	\path[-{>[scale=2.5,length=2,width=4]},line width=0.4pt] (dbored) [dashed] edge node [above] {} (creativeactive.east);
	\path[-{>[scale=2.5,length=2,width=4]},line width=0.4pt] (isbored) [dashed] edge node [left] {yes} (boredom1);
	\path[-] (boredom1) [dashed] edge node {} (d1);
	\path[-{>[scale=2.5,length=2,width=4]},line width=0.4pt] (d1) [dashed] edge node {} (exec1);
	\path[-{>[scale=2.5,length=2,width=4]},line width=0.4pt] (boredomactive) edge node [left] {no} (creativeactive);
	\path[-{>[scale=2.5,length=2,width=4]},line width=0.4pt] (creativeactive.west) edge node [above] {no} (exec3.south);
	\path[-{>[scale=2.5,length=2,width=4]},line width=0.4pt] (exec3) edge node {} (exec4);
	\path[-{>[scale=2.5,length=2,width=4]},line width=0.4pt] (exec4) edge node {} (welltrained);
	\path[-{>[scale=2.5,length=2,width=4]},line width=0.4pt] (welltrained) edge node [right] {yes} (finished1);
	\path[-{>[scale=2.5,length=2,width=4]},line width=0.4pt] (welltrained.east) edge node [right] {no} (d2);
	\path[-{>[scale=2.5,length=2,width=4]},line width=0.4pt] (d2) edge node {} (exec1);
	\path[-] (creativeactive.south) [dashed] edge node {} (d3);
	\path[-] (d3) [dashed] edge node [above] {yes} (d4);
	\path[-{>[scale=2.5,length=2,width=4]},line width=0.4pt] (d4) [dashed] edge node {} (creative1);
	\path[-{>[scale=2.5,length=2,width=4]},line width=0.4pt] (creative1) [dashed] edge node {} (exec3);
	
	\path[-{>[scale=2.5,length=2,width=4]},line width=0.4pt] (exec3) [dashed] edge node {} (env1);
	\path[-{>[scale=2.5,length=2,width=4]},line width=0.4pt] (env1) [dashed] edge node {} (env2);
	\path[-{>[scale=2.5,length=2,width=4]},line width=0.4pt] (env2) [dashed] edge node {} (exec4);
	
\end{tikzpicture} 
}
\caption{Flow chart of the skill training procedure.
A novel skill is trained by showing a new basic behaviour.
The robot extends the domain of applicability by playing the object, i.e.
by performing a random walk through the network shown in Fig. \ref{fig:ecmplaying}.
The solid lines indicate the behaviour of the basic approach \cite{Hangl-2016-IROS}.
The shaded areas and dashed lines show the proposed extensions, i.e.
\emph{training of an environment model}, \emph{boredom} and \emph{creative skill generation}.
}
\label{fig:execlearnpath}
\end{figure}
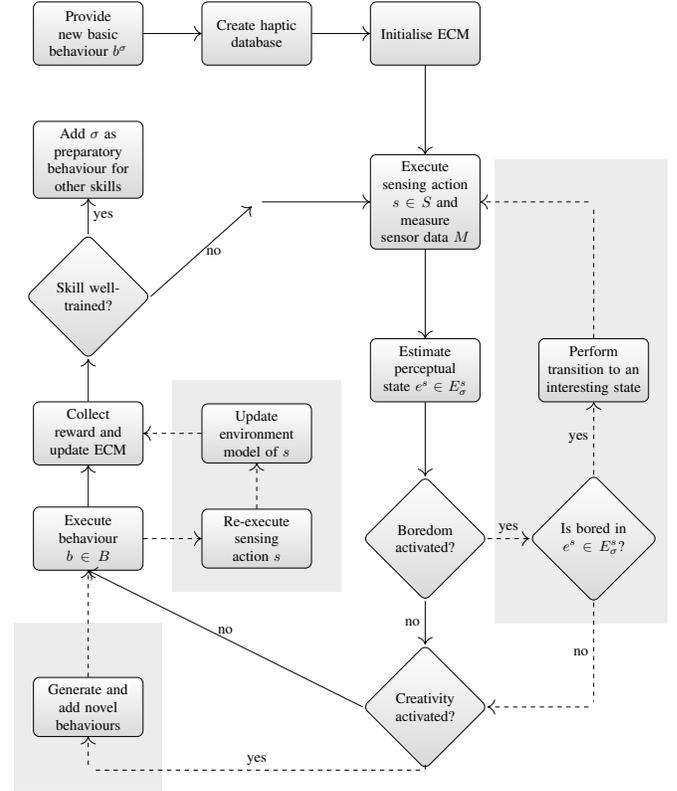
The domain of applicability of a skill $\sigma$ is extended
by running the PS as described in section~\ref{sec:projectivesimulation}
on the ECM in Fig.~\ref{fig:ecmplaying}.
The robot collects reward after each rollout and updates the
transition probabilities accordingly.
Skills are added as preparatory behaviours of other skills
as soon as they are \emph{well-trained}, i.e. the average
reward $\bar{r}$ over the last $t_{\mathrm{thresh}}$ rollouts
reaches a threshold $\bar{r} \geq r_{\mathrm{thresh}}$.
This enables the robot to create increasingly complex skill hierarchies.
The complete training procedure of a skill $\sigma$ is shown in
Fig.~\ref{fig:execlearnpath}.
Only the non-shaded parts and solid transitions are
available in this basic version.
\subsection{Properties and extensions}
A strong advantage is that state-of-the-art research on object manipulation can be embedded
by adding the controllers to the set of behaviours.
Algorithms for specific problems (e.g. grasping, pushing
\cite{Krivic-2016-CASE, pushforgrasping, mulling2013learning, 591646, whitney1987historical})
can be re-used in a bigger framework that orchestrates their interaction.

In the basic version the state space is comparatively small,
which enables the robot to learn skills without an environment model.
Further, the robot to learn fast while still preserving the
ability to learn quite complex skills autonomously.
However, the lack an environment model can be both an advantage and a disadvantage.
Testing a hypothesis directly on the environment enables the robot
to apply behaviours outside of the intended context (e.g. a book flipping
behaviour might be used to open a box~\cite{Hangl-2016-IROS}).
This is hard to achieve with model-based approaches if the modeled
domain of a behaviour cannot properly represent the relevant information.
On the other hand, the lack of reasoning abilities limits the learning speed and the
complexity of solvable problems.
We overcome this problem by additionally learning an environment model
from information acquired during playing.
The robot learns a distribution of the effects of behaviours
on given perceptual states by re-estimating the state after execution.
We use the environment model for two purposes: \emph{active learning}
and \emph{creative generation of novel preparatory behaviours}.

The basic version intrinsically assumes that all
required preparatory behaviours are available.
This constitutes a strong prior and limits the degree of autnomoy.
We weaken this requirement by allowing the robot
to creatively generate potentially useful combinations of behaviours.
These are made avaible for the playing system which tries them out.
Further, experiments showed that the learning speed was decreased
by performing rollouts in situations that were already solved before.
We use the environment model to implement \emph{active learning}.
Instead of asking a supervisor to prepare interesting situations,
the robot prepares them by itself.
\section{Learning an Environment Model}
\label{sec:envmodel}
\begin{figure*}[t!]
\centering
  \scalebox{0.65} {
\begin{tikzpicture}[every node/.style = {shape=rectangle, rounded corners},>=stealth',bend angle=45, auto, font=\fontsize{11}{20}\selectfont]

  \tikzstyle{place}=[circle,thick,draw=blue!75,fill=blue!20,minimum size=6mm]

  \begin{scope}
  
  	\draw(0,1) node[ellipse, minimum height=1.3cm, minimum width=2.5cm,draw] (c1) {$e^{s}_{1} b_1$};
  	\draw(3,1) node[ellipse, minimum height=1.3cm, minimum width=2.5cm,draw] (c2) {$e^{s}_{1} b_2$};
  	\draw(6,1) node[ellipse, minimum height=1.3cm, minimum width=2.5cm,draw] (c3) {$\dots$};
  	\draw(9,1) node[ellipse, minimum height=1.3cm, minimum width=2.5cm,draw] (c4) {$e^{s}_{1} b_J$};
  	\draw(12,1) node[ellipse, minimum height=1.3cm, minimum width=2.5cm,draw] (c5) {$\dots$};
  	\draw(15,1) node[ellipse, minimum height=1.3cm, minimum width=2.5cm,draw] (c6) {$e^{s}_{L_{\sigma,s}} b_1$};
  	\draw(18,1) node[ellipse, minimum height=1.3cm, minimum width=2.5cm,draw] (c7) {$\dots$};
  	\draw(21,1) node[ellipse, minimum height=1.3cm, minimum width=2.5cm,draw] (c8) {$e^{s}_{L_{\sigma,s}} b_{J}$};
  	
  	\draw(6, -2.5) node[circle, minimum size=1.2cm, inner sep=0pt,draw] (e1) {$e^{s}_1$};
  	\draw(9, -2.5) node[circle, minimum size=1.2cm, inner sep=0pt,draw] (e2) {$e^{s}_2$};
  	\draw(12, -2.5) node[circle, minimum size=1.2cm, inner sep=0pt,draw] (e3) {$\dots$};
  	\draw(15, -2.5) node[circle, minimum size=1.2cm, inner sep=0pt,draw] (e4) {$e^{s}_{L_{\sigma,s}}$};
  	
  	\path[->] (c1.south) edge node {} (e1.north);
  	\path[->] (c1.south) edge node {} (e2.north);
  	\path[->] (c1.south) edge node {} (e3.north);
  	\path[->] (c1.south) edge node {} (e4.north);
  	
  	\path[->] (c2.south) edge node {} (e1.north);
  	\path[->] (c2.south) edge node {} (e2.north);
  	\path[->] (c2.south) edge node {} (e3.north);
  	\path[->] (c2.south) edge node {} (e4.north);
  	
  	\path[->] (c3.south) edge node {} (e1.north);
  	\path[->] (c3.south) edge node {} (e2.north);
  	\path[->] (c3.south) edge node {} (e3.north);
  	\path[->] (c3.south) edge node {} (e4.north);
  	
  	\path[->] (c4.south) edge node {} (e1.north);
  	\path[->] (c4.south) edge node {} (e2.north);
  	\path[->] (c4.south) edge node {} (e3.north);
  	\path[->] (c4.south) edge node {} (e4.north);
  	
  	\path[->] (c5.south) edge node {} (e1.north);
  	\path[->] (c5.south) edge node {} (e2.north);
  	\path[->] (c5.south) edge node {} (e3.north);
  	\path[->] (c5.south) edge node {} (e4.north);
  	
  	\path[->] (c6.south) edge node {} (e1.north);
  	\path[->] (c6.south) edge node {} (e2.north);
  	\path[->] (c6.south) edge node {} (e3.north);
  	\path[->] (c6.south) edge node {} (e4.north);
  	
  	\path[->] (c7.south) edge node {} (e1.north);
  	\path[->] (c7.south) edge node {} (e2.north);
  	\path[->] (c7.south) edge node {} (e3.north);
  	\path[->] (c7.south) edge node {} (e4.north);
  	
  	\path[->] (c8.south) edge node {} (e1.north);
  	\path[->] (c8.south) edge node {} (e2.north);
  	\path[->] (c8.south) edge node {} (e3.north);
  	\path[->] (c8.south) edge node {} (e4.north);
    
  \end{scope}
\end{tikzpicture}
}
\caption{ECM for the environment model under the sensing action $s$.
It reflects the probability of state transitions $e^{s} \xrightarrow{b} e'^{s}$.
The environment model is trained by adding a sensing step after executing the preparatory behaviour $b \in B$.
}
\label{fig:envmodel}
\end{figure*}
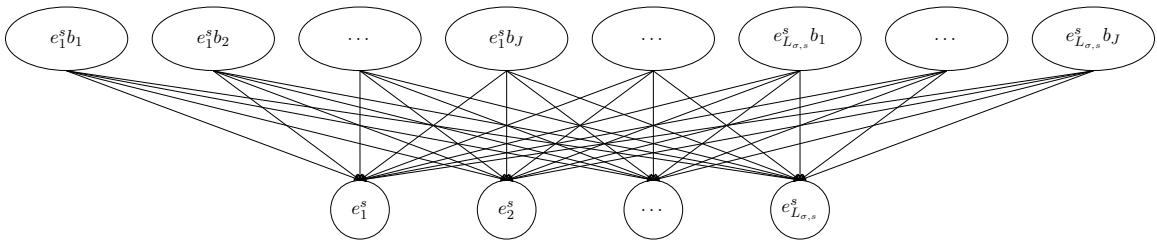
The \emph{environment model} predicts the effect, i.e. the resulting
perceptual state, of a behaviour on a given perceptual state.
An environment model is the probability distribution
$p\left(e'^{s} \, | \, e^{s}, \, b, \, \sigma \right)$ where $e^{s}, e'^{s} \in E^s_{\sigma}$
are perceptual states of the sensing action $s \in S$ for a skill $\sigma$,
and $b \in B$ is a behaviour.
It denotes the probablity of the transition $e^{s} \xrightarrow{b} e'^{s}$.
The required training data is acquired by re-executing the sensing action $s$
after applying the behaviour $b$, c.f. shaded center part in Fig.~\ref{fig:execlearnpath}.
Given a playing sequence $\sigma \xrightarrow{s} e^{s} \xrightarrow{b} e'^{s}$
(c.f. Fig.~\ref{fig:ecmplaying}) the effect can be observed by re-executing $s$ with
\begin{equation}
\sigma \xrightarrow{s} e^{s} \xrightarrow{b} e'^{s} \xrightarrow{s} e'^{s}
\label{equ:envdetpath}
\end{equation}
The assumptions in equations \ref{equ:assumptionsense}~-~\ref{equ:assumptionprep}
forbid to additionally execute other sensing actions $s' \in S$ without influencing
the playing based method.
This limitation prevents the robot
from learning more complex environment models as done in related work
\cite{oudeyer-iac, oudeyer1-riac, Ugur-2016-TCDS, stoytchev2004incorporating,
barto2004intrinsically, schembri2007evolution}, e.g. capturing transitions
between perceptual states of different sensing actions.
However, the purpose of the environment model is not to perform precise plans
but to feed the core playing component with new \emph{ideas}.

We represent the distribution $p\left(e'^{s} \, | \, e^{s}, \, b, \, \sigma \right)$
by another ECM for each skill~-~sensing action pair $(\sigma, s)$
as shown in Fig. \ref{fig:envmodel}.
The percept clips consist of pairs $\left(e^s, b\right)$ of perceptual states
$e^s \in E^s_{\sigma}$ and preparatory behaviours $b \in B$.
The target clips are the possible resulting states $e'^s \in E^s_{\sigma}$.
The environment model is initialised with uniform weights
$h^{\mathrm{env}}_{(e^s, b) \rightarrow e'^s} = 1$.
When a skill $\sigma$ is executed using the path in equation \ref{equ:envdetpath},
a reward of $r^{\mathrm{env}} \in \mathbb{R}^+$ is given for
the transition
\begin{equation}
	\left( e^{s}, b \right) \rightarrow e'^{s}
\end{equation}
and the weights are updated accordingly, c.f. equation \ref{equ:pslearn}.
When a novel preparatory behaviour $b_{K + 1}$ is available
for playing, e.g. a skill is well-trained and is
added as a preparatory behaviour,
it is included into the environment models for each skill~-~sensing action pair
$(\sigma, s)$ by adding clips $(e^s, b_{K + 1})$ for all states $e^s \in E^s_{\sigma}$
and by connecting them to all $e'^s \in E^s_{\sigma}$ with the uniform
initial weight $h_{\text{init}}^{\mathrm{env}} = 1$.

We employ a practical restriction on the scope of the environment model.
The additional sensing action execution  is only done if the grasp outcome
of the seleced preparatory behaviour and the grasp requirement of the
the sensing action match, e.g. if the preparatory behaviour grasps
the object, but the sensing action was \emph{sliding}, re-execution of the
sensing action would destroy the grasp and is not done.
\section{Autonomous Active Learning}
\label{sec:activelearning}
In the basic version an optimal selection of
observed perceptual states is required in order to learn the correct behaviour
in all possible states, i.e. in a semi-supervised setting a
human supervisor should mainly prepare unsolved perceptual states.
This would require the supervisor to have knowledge about the method itself
and about the semantics of perceptual states,
which is an undesirable property.
Instead, we propose to equip the robot with the ability to reject
perceptual states in which the skill is well-trained already.
In an autonomous setting, this is not sufficient as
it would just stall the playing.
The robot has to prepare a more interesting state autonomously.
We propose to plan state transitions by using the environment model
in order to reach states which (i) are \emph{interesting} and
(ii) can be prepared with high confidence.
We can draw a loose connection to human behaviour.
In that spirit, we call the rejection of well-known states \emph{boredom}.
\subsection{Boredom}
\label{sec:boredom}
The robot may be \emph{bored} in a given perceptual state, if it is
\emph{confident} about the task solution, i.e. if the distribution of
which preparatory behaviour to select is highly concentrated.
In general, every function reflecting uncertainty can be used.
We use the normalised Shannon entropy to measure the confidence 
in a perceptual state $e \in E^s_{\sigma}$, given by
\begin{equation}
\begin{split}
\hat{H}_{e} = \frac{H \left( b \, | \, e \right)}{H_{\mathrm{max}}} = -\frac{
\sum_{b' \in B}{ p\left( b = b' \, | \, e \right) \log_2{p\left( b = b' \, | \, e \right)}}
}{\log_2 J}
\end{split}
\end{equation}
where $J$ is the number of preparatory behaviours.
If the entropy is high, the robot either has not learned anything yet
(and therefore all the transition weights are close to uniform) or it
observes the degenerate case that all preparatory behaviours
deliver (un)successful execution (in which case there is nothing to
learn at all).
If the entropy is low, few transitions are strong,
i.e. the robot knows well how to handle this situation.
We use the normalised entropy to define the probability
of being bored in a state $e \in E^s_{\sigma}$ with
\begin{equation}
p \left(\mathrm{bored} = true \, | \, e \right) = 1 - \beta \hat{H}_{e}
\label{equ:boredomprob}
\end{equation}
The constant $\beta \in [ 0, 1 ]$ defines
how \emph{immune} the agent is to boredom.
The robot samples according to $p \left(\mathrm{bored}\, | \, e \right)$
and decides on whether to refuse the execution.
\subsection{Transition Confidence}
If the robot is bored in a perceptual state $e' \in E^s_{\sigma}$,
it autonomously tries to prepare a more interesting state $\hat{e} \in E^s_{\sigma}$.
This requires the notion of a \emph{transition confidence} for which
the environment model can be used.
We aim to select behaviours conservatively
which allows the robot to be certain about the effect of the transition.
We do not use the probability of reaching one state from another directly,
but use a measure considering the complete distribution $p(e \, | \, e', b)$.
By maximising the normalised Shannon entropy, we favour deterministic transitions.
For each state-action pair $(e', b)$ in Fig.~\ref{fig:envmodel}
we define the transition confidence $\nu^{s}_{e' b}$ by
\begin{equation}
\nu^{s}_{e' b} = 1 - \frac{H \left( e \, | \, (e', b) \right)}{H_{\mathrm{max}}} =
1 - \frac{H \left( e \, | \, (e', b) \right)}{\log_2{L_{\sigma, s}}}
\label{equ:singlesteptransition}
\end{equation}
where $e' \in E^s_{\sigma}$, $b \in B$, and $L_{\sigma, s}$ is the number
of perceptual states under the sensing action $s \in S$, i. e.
the number of children of the clip $\left( e', b \right)$.
In contrast to the entropy computed in section \ref{sec:boredom},
the transition confidence is computed on the
environment model, c.f. Fig. \ref{fig:envmodel}.
The \emph{successor function} $\text{su}(e, b)$ returns the most likely
resulting outcome of executing behaviour $b$ in a perceptual state $e \in E^s_{\sigma}$
and is defined by
\begin{equation}
	\text{su}(e, b) = \argmax_{e'}{p_{(e, b) \rightarrow e'}}
\label{eq:greedysuccessor}
\end{equation}
In practice, single state transitions are not sufficient.
For paths $e = e^{s}_{1} \xrightarrow{b_{1}}
e^{s}_{2} = \text{su}(e^{s}_{1}, b_1) \xrightarrow{b_{2}} \dots \xrightarrow{b_{L - 1}} \text{su}(e^{s}_{n_{L - 1}}, b_{L - 1})
= e^{s}_{L} = e'$ of length $L$ we define the transition confidence with
\begin{equation}
\nu^{s}_{e \mathbf{b}} = \prod_{l = 1}^{L - 1}{\nu^{s}_{e_l b_l}}
\label{eq:multipathconf}
\end{equation}
where the vector $\mathbf{b} = \left( b_1, b_2, \dots, b_{L - 1} \right)$
denotes the sequence of behaviours.
This is equivalent to a greedy policy, which provides a more
conservative estimate of the transition confidence and eliminates
consideration of transitions that could occur by pure chance.
A positive side effect is the efficient computation of equation \ref{eq:multipathconf}.
Only the confidence of the most likely path is computed instead
of iterating over all possible paths.
The path $\mathbf{b}$ is a behaviour itself and the successor
is given by $\text{su}(e, \mathbf{b}) = \text{su}(e^s_{n_{L - 1}}, b_{L - 1})$.
\subsection{Active Learning}
If the robot encounters a boring state $e \in E^s_{\sigma}$,
the goal is to prepare the most interesting state that can
acutally be produced.
We maximise the \emph{desirability function} given by
\begin{equation}
\left( \mathbf{b}, L \right) = \argmax_{\mathbf{b}, L}{\left[ \hat{H}_{
\text{su}(e, \mathbf{b})} \nu_{e\mathbf{b}} + \frac{\epsilon}{\mathrm{cost}\left( \mathbf{b} \right)} \right]}
\label{equ:excitementfunction}
\end{equation}
where $\hat{H}_{\text{su}(e, \mathbf{b})}$ is the entropy of the expected
final state and $\nu_{e\mathbf{b}}$ is the confidence of reaching the final
state by the path $\mathbf{b}$.
The \emph{balancing factor} $\epsilon$ defines the relative importance of the
desirability and the path cost.
The path cost $\mathrm{cost}\left( \mathbf{b} \right)$
can be defined by the length of the path $L$, i.e. penalising long paths,
or, for instance, by the average execution time of $\mathbf{b}$.
Equation~\ref{equ:excitementfunction}
balances between searching for an interesting state while
making sure that it is reachable.
In practice it can be optimised by enumerating all
paths of reasonable length, e.g. $L < L_{\mathrm{max}}$,
with typical values of $L_{\mathrm{max}} \leq 4$.

The basic method is extended by sampling from the boredom distribution
after the state estimation.
If the robot is bored, it optimises the desirability function and
executes the transition to a more interesting state.
This is followed by restarting the skill execution with
boredom turned off in order to avoid boredom loops,
c.f. right shaded box in Fig.~\ref{fig:execlearnpath}.
\section{Creative Behaviour Generation}
\label{sec:creativity}
\begin{figure*}[t!]
  \centering
  \subfloat[The behaviour $b_{J + 1}$ is added the initial weight $h_{\text{init}}$ (dashed lines) to all perceptual
states except for the current state $e^s$.
In this case it is added with a higher weight proportional to the curiosity score (solid line),
c.f. equation \ref{equ:creativeaddprobecm}.
The connections to all other preparatory behaviours are ommitted in this
figure.]{\label{fig:creativeinsertecm}
  \resizebox{0.55\textwidth}{!}{%
  	\begin{tikzpicture}[every node/.style = {shape=rectangle, rounded corners},>=stealth',bend angle=45, auto]

  	\draw(2.25,0.1) node[rectangle, minimum width=8cm, minimum height=4.3cm,draw] (sensingbox1) {};
  	\node[text width=6cm,align=left, font=\fontsize{14}{20}\selectfont] at (1.5,1.8) {Sensing action $s$};
  	\draw(2.25,1.5) node[circle, minimum size=1cm,draw, font=\fontsize{15}{20}\selectfont] (s1) {$s$};
  
	\draw(2.25,-0.75) node[rectangle, minimum width=7cm, minimum height=2cm,draw] (statebox1) {};
	\node[text width=6cm,align=left, font=\fontsize{14}{20}\selectfont] at (2,-0.1) {Perceptual states of $s$};
    \draw(0.0,-1.0) node[circle, minimum size=1cm,draw,dashed, font=\fontsize{15}{20}\selectfont] (e11) {$e_1^{s}$};
    \draw(1.5,-1.0) node[circle, minimum size=1cm,draw, font=\fontsize{15}{20}\selectfont] (e12) {$e^{s}$};
    \draw(3.0,-1.0) node[circle, minimum size=1cm,draw,dashed, font=\fontsize{15}{20}\selectfont] (e1dot) {$e_{\dots}^{s}$};
    \draw(4.5,-1.0) node[circle, minimum size=1cm,draw,dashed, font=\fontsize{15}{20}\selectfont] (e1L1) {$e_{L_{\sigma,s}}^{s}$};
    
    \draw(8.75,0.1) node[rectangle, minimum width=3cm, minimum height=4.3cm,draw,dashed, font=\fontsize{14}{20}\selectfont] (sensingbox2) {%
    Sensing action $s_{\dots}$};
    
    \draw(13.0,0.1) node[rectangle, minimum width=3cm, minimum height=4.3cm,draw,dashed, font=\fontsize{14}{20}\selectfont] (sensingbox3) {%
    Sensing action $s_N$};
    
    \draw(5.0,-3.5) node[rectangle, minimum width=6cm, minimum height=1.5cm,draw, font=\fontsize{14}{20}\selectfont] (a1) {%
    \begin{varwidth}{6cm}
    $b_{J + 1} \equiv$ Novel preparatory behaviour
    \end{varwidth}
    };
    
    \path[->] (s1.south) edge [line width=0.2mm] node {} (e11.north);
    \path[->] (s1.south) edge [line width=0.2mm] node {} (e12.north);
    \path[->] (s1.south) edge [line width=0.2mm] node {} (e1dot.north);
    \path[->] (s1.south) edge [line width=0.2mm] node {} (e1L1.north);
    
    \path[->] (e11.south) edge [line width=0.2mm,dashed] node {} (a1.north);
    \path[->] (e12.south) edge [line width=1.5] node {} (a1.north);
    \path[->] (e1dot.south) edge [line width=0.2mm,dashed] node {} (a1.north);
    \path[->] (e1L1.south) edge [line width=0.2mm,dashed] node {} (a1.north);
    \path[->] (sensingbox2.south) edge [line width=0.2mm,dashed] node {} (a1.north);
    \path[->] (sensingbox3.south) edge [line width=0.2mm,dashed] node {} (a1.north);
  
\end{tikzpicture}
  }
  } \quad
  \subfloat[All pairs $(e, b_{J + 1})$ of perceptual states $e \in E^s_{\sigma}$ and the behaviour $b_{J + 1} = \mathbf{b}$ are added.
The weights are chosen according to equation \ref{equ:creativeaddprobenv}
(case 1: solid line, case 2: dashed lines).
Note that case 1 only applies for the currently used sending action $s \in S$.
]{\label{fig:creativeinsertenv}
  \resizebox{0.4\textwidth}{!}{%
  	\begin{tikzpicture}[every node/.style = {shape=rectangle, rounded corners},>=stealth',bend angle=45, auto, font=\fontsize{15}{20}\selectfont]

  \tikzstyle{place}=[circle,thick,draw=blue!75,fill=blue!20,minimum size=8mm]

  \begin{scope}
  
  	\draw(4.5,2) node[ellipse, minimum height=1.3cm, minimum width=2.5cm,draw, dashed] (c4) {$e^s_1 \mathbf{b}$};
  	\draw(7.5,2) node[ellipse, minimum height=1.3cm, minimum width=2.5cm,draw, dashed] (c5) {$\dots$};
  	\draw(10.5,2) node[ellipse, minimum height=1.3cm, minimum width=2.5cm,draw] (c6) {$e^s \mathbf{b}$};
  	\draw(13.5,2) node[ellipse, minimum height=1.3cm, minimum width=2.5cm,draw, dashed] (c7) {$\dots$};
  	\draw(16.5,2) node[ellipse, minimum height=1.3cm, minimum width=2.5cm,draw, dashed] (c8) {$e^s_{N_s} \mathbf{b}$};
  	
  	\node[text width=5cm,align=left] at (8,-4.3) {};
  	
  	\draw(6, -2.5) node[circle, minimum size=1.7cm, inner sep=0pt,draw] (e1) {$e^s_{1}$};
  	\draw(9, -2.5) node[circle, minimum size=1.7cm, inner sep=0pt,draw] (e2) {
  	$\text{su}(e^s, b)$};
  	\draw(12, -2.5) node[circle, minimum size=1.7cm, inner sep=0pt,draw] (e3) {$\dots$};
  	\draw(15, -2.5) node[circle, minimum size=1.7cm, inner sep=0pt,draw] (e4) {$e_{L_{\sigma, s}}$};
  	
	\path[->] (c4.south) edge [dashed] node {} (e1.north);
  	\path[->] (c4.south) edge [dashed] node {} (e2.north);
  	\path[->] (c4.south) edge [dashed] node {} (e3.north);
  	\path[->] (c4.south) edge [dashed] node {} (e4.north);  	
  	
  	\path[->] (c5.south) edge [dashed] node {} (e1.north);
  	\path[->] (c5.south) edge [dashed] node {} (e2.north);
  	\path[->] (c5.south) edge [dashed] node {} (e3.north);
  	\path[->] (c5.south) edge [dashed] node {} (e4.north);
  	
  	\path[->] (c6.south) edge [dashed] node {} (e1.north);
  	\path[->] (c6.south) edge [dashed] node {} (e3.north);
  	\path[->] (c6.south) edge [dashed] node {} (e4.north);
  	
  	\path[->] (c7.south) edge [dashed] node {} (e1.north);
  	\path[->] (c7.south) edge [dashed] node {} (e2.north);
  	\path[->] (c7.south) edge [dashed] node {} (e3.north);
  	\path[->] (c7.south) edge [dashed] node {} (e4.north);
  	
  	\path[->] (c8.south) edge [dashed] node {} (e1.north);
  	\path[->] (c8.south) edge [dashed] node {} (e2.north);
  	\path[->] (c8.south) edge [dashed] node {} (e3.north);
  	\path[->] (c8.south) edge [dashed] node {} (e4.north);
  	
  	\path[->,line width = 1.5] (c6.south) edge node [] {} (e2.north);
    
  \end{scope}
\end{tikzpicture}
  }
  }
\caption{Insertion of the novel compound behaviour $b_{J + 1} = \mathbf{b}$, creatively
  generated in the current perceptual state $e^s \in E^s_{\sigma}$, to the
  playing ECM of skill $\sigma$ (left), c.f. Fig. \ref{fig:ecmplaying}, and the environment
  models of the used sensing action $s \in S$ (right) respectively.}
\label{fig:insertbehaviour}
\end{figure*}
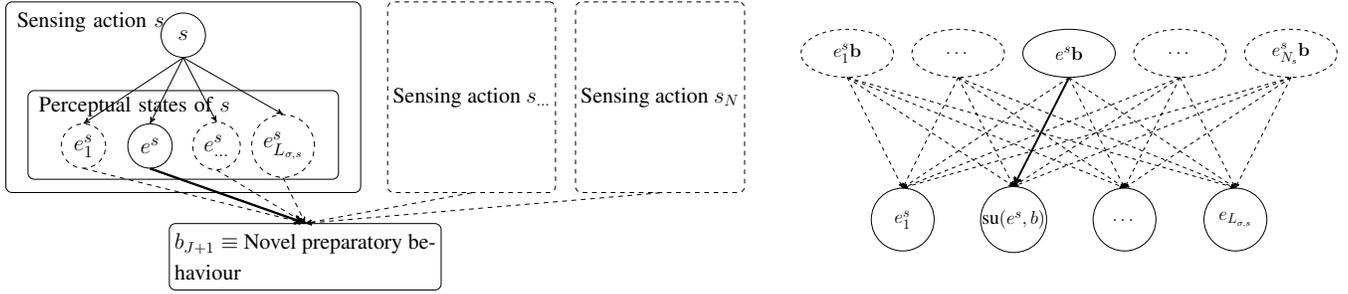
In many cases, the required preparatory behaviour
is a combination of other available behaviours,
e.g. \emph{rotate $180^{\circ}$} $\equiv$
\emph{rotate $90^{\circ}$} + \emph{rotate $90^{\circ}$}.
Without using some sort of intelligent reasoning, the space
of concatenated behaviours explodes and becomes intractable.
However, any sequence of behaviours that transfers the current unsolved state
to a target state, i.e. a state which does not require any preparation,
is potentially useful as a compound behaviour itself.
Sequences can be generated by planning transitions to target states.
If the robot is bored, it uses active learning, if not,
the situation is not solved yet and novel compound behaviours
might be useful.

A perceptual state $e_{\emptyset} \in E^s_{\sigma}$ is a target state if
the transition with the highest probability in the playing ECM
(Fig.~\ref{fig:ecmplaying}) leads to the
\emph{void}-behaviour with $p_{e_{\emptyset} \rightarrow b_{\emptyset}}$.
If there exists a path
$e^s \xrightarrow{\mathbf{b}} e_{\emptyset}$
from the current perceptual state $e^s \in E^s_{\sigma}$ to a target state $e_{\emptyset}$,
the sequence $\mathbf{b} = \left( b_1, \dots, b_L \right)$
is a candidate for a novel behaviour.
The robot is \emph{curious} about trying out the novel
compound behaviour $\mathbf{b}$,
if the transition confidence $\nu_{e^s \mathbf{b}}$, with
$\text{su}(e^s, \mathbf{b}) = e_{\emptyset}$, and
the probability $p_{e_{\emptyset} \rightarrow b_{\emptyset}}$
of the state actually being a real target state are both high.
This is measured by the \emph{curiosity score} of the compound behaviour
given by
\begin{equation}
	\text{cu}(e^s, \mathbf{b}) = \nu_{e^s \mathbf{b}} \, p_{\text{su}(e^s, \mathbf{b}) \rightarrow b_{\emptyset}}
\end{equation}
The factor $p_{\text{su}(e^s, \mathbf{b}) \rightarrow b_{\emptyset}}$
reduces the score in case the state $e_{\emptyset}$ is a target state
with low probability.
This can happen if in previous rollouts all other behaviours
were executed and were punished.
We use a probability instead of a confidence value to allow
creativity even in early stages where a target
state was not identified with a high probability.

The compound behaviour with the highest score is added
as novel behaviour $b_{J + 1} = \mathbf{b}$ with the
probability given by squashing the curiosity score into
the interval $[0, 1]$ with
\begin{equation}
p \left( \text{add } b_{J + 1} = \mathbf{b} \, | \, e^s \right) =
\text{sig} \left[ \gamma \, \text{cu}\left( e^s, \mathbf{b} \right) + \delta \right]
\label{equ:creativeaddprob}
\end{equation}
where $\text{sig}$ is the logistic sigmoid.
The parameters $\gamma, \delta$ define how conservatively
novel behaviour proposals are created.
The novel behaviour $b_{J + 1}$ is added as preparatory
behaviour for all perceptual states under the current skill
$\sigma$ with the weights
\begin{equation}
\begin{split}
h_{e \rightarrow b_{J + 1}} = 
\begin{cases}
h_{\text{init}} \left[ 1 + \text{cu}\left( e^s, b_{J + 1} \right) \right] &, \text{ if } e = e^s\\
h_{\text{init}} &, \text{ else}
\end{cases}
\end{split}
\label{equ:creativeaddprobecm}
\end{equation}
It is added with at least the initial weight
$h_{\text{init}}$, but increased proportional to the curiosity score
for the current perceptual state $e^s \in E^s_{\sigma}$,
c.f. Fig.~\ref{fig:creativeinsertecm}.
The novel behaviour is also inserted to the environment model of
all sensing actions $s \in S$.
For each perceptual state $e \in E^s_{\sigma}$,
a clip $(e, b)$ is added and connected to the clips $e' \in E^s_{\sigma}$
in second layer with the weights
\begin{equation}
\begin{split}
h_{ \left( e, b \right) \rightarrow e'} = 
\begin{cases}
h_{\text{min}}\left( b \right) &, \text{ if }  e = e^s, \, e' = \text{su}(e^s, b), \, b = b_{J + 1} \\
h^{\mathrm{env}}_{\mathrm{init}} &, \text{ else}
\end{cases}
\end{split}
\label{equ:creativeaddprobenv}
\end{equation}
where $h_{\text{min}}\left( b_{J + 1} \right) = h_{\text{min}}\left( \mathbf{b} \right)$ is the minimum
transition value on the path $\mathbf{b}$ through the environment model,
following the idea that a chain is only as strong as its weakest link.
The weights of all other transitions are set to the initial weight
$h^{\mathrm{env}}_{\mathrm{init}}$, c.f. Fig.~\ref{fig:creativeinsertenv}.
\section{Experiments}
We evaluate our method using a mix of simulated and real-world experiments.
Our real-world experiments cover a wide range of skills to show the expressive power.
We show how skill hierarchies are created within our framework.
Success statistics of the single components (sensing accuracy,
success rate of preparatory behaviours, success rate of basic
behaviours) were used to assess the convergence behaviour by simulation.
Table \ref{tab:paramvalues} lists the used parameter values.
We execute all skills and behaviours in impedance mode in order to prevent
damage to the robot.
Further, executed behaviours are stopped if a maximum force is exceeded.
This is a key aspect for model-free playing, which enables the robot
to try out arbitrary behaviours in arbitrary tasks.
\begin{table}[h]
\centering
\begin{tabular}{ l l l }
Parameter & Name & Values\\
\hline
Skill success reward & $r(\text{success})$ & 1000 \\
Skill failure punishment & $r(\text{failure})$ & -30 \\
PS forgetting factor & $\zeta$ & 0 \\
Environment model reward & $r^{\text{env}}$ & 10 \\
Skill ECM intial weight & $h_{\text{init}}$ & 200 \\
Environment model ECM intial weight & $h_{\text{init}}^{\text{env}}$ & 1 \\
Stretching factor & $\alpha$ & 25 \\
Boredom immunity & $\beta$ & 0.8 \\
Squashing scale & $\gamma$ & 0.1 \\
Squashing shift & $\delta$ & 0.95 \\
Balancing factor & $\epsilon$ & 0.1 \\
Maximum creativity path length & $L_{\text{max}}$ & 4 \\
\hline
\\
\end{tabular}
\caption{List of free parameters and values used}
\label{tab:paramvalues}
\end{table}
\subsection{Experimental Setup}
\label{sec:expsetup}
\begin{figure}[t!]
 \centering
	 \includegraphics[width=0.5\textwidth]{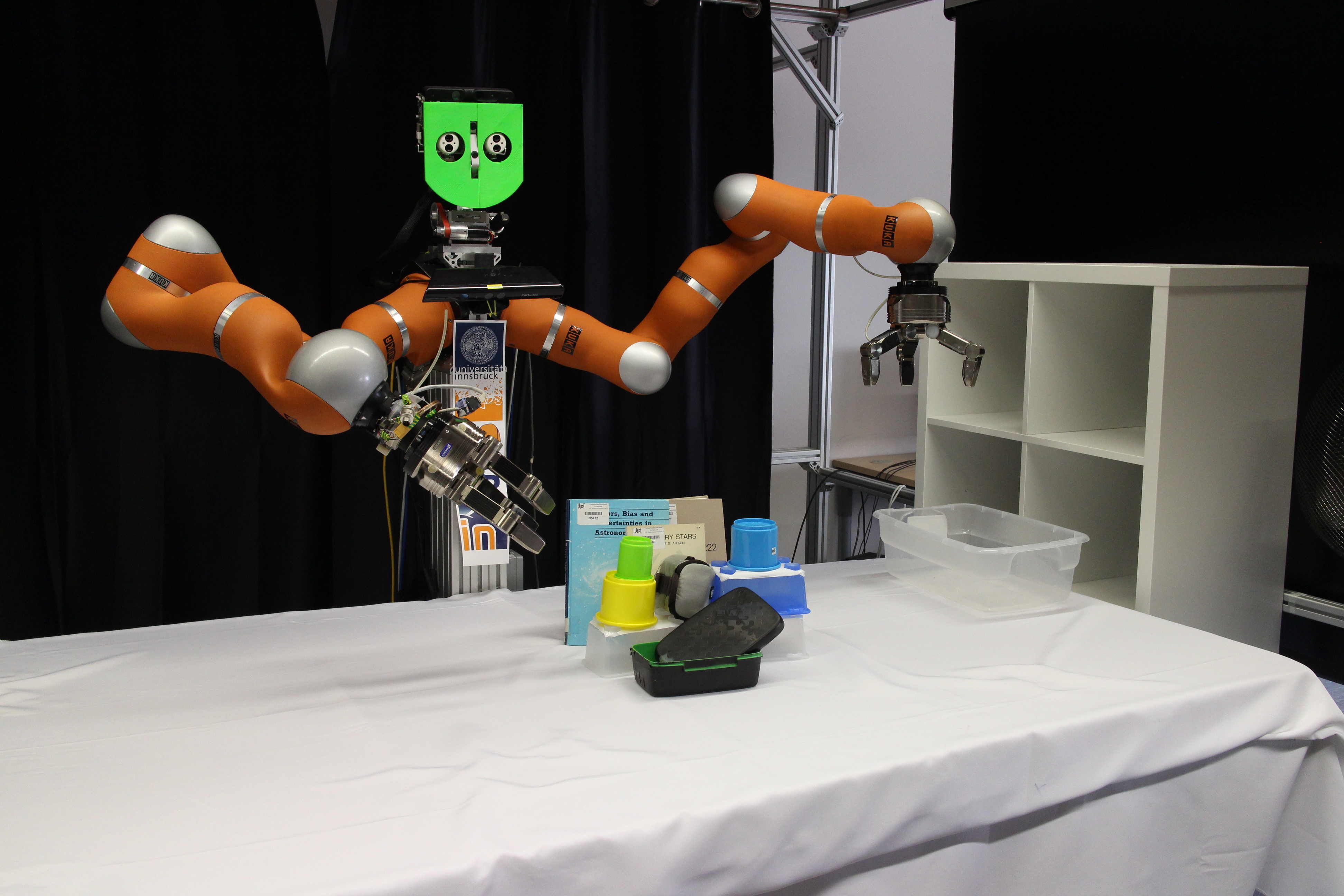}
  \caption{Robot setting and used objects.
  The hardware included a Kinect, two KUKA LWR 4+ and
  two Schunk SDH grippers.
  The objects used for the trained tasks were books of different dimensions
  and cover types, an IKEA shelf and boxes, and selected objects of the YCB object and model set \cite{calli2015ycb}.}
  \label{fig:setting}
\end{figure}

The robot setting is shown in Fig.~\ref{fig:setting}.
For object detection a Kinect mounted above the robot is used.
All required components and behaviours are implemented
with the \emph{kukadu} robotics framework\footnote{\url{https://github.com/shangl/kukadu}}.
The perceptual states are estimated from joint positions, Cartesian end-effector positions,
joint forces and Cartesian end-effector forces / torques.
Objects are localised by removing the table surface from the point
cloud and fitting a box by using PCL.
Four controllers implement the available preparatory behaviours:
\begin{itemize}
	\item \emph{Void behaviour}: The robot does nothing.
	\item \emph{Rotation}: The object is rotated by
	a circular finger movement around the object's center.
	The controller can be parametrised with the approximate rotation angle.
	\item \emph{Flip}: The object is squeezed between the hands
	and one hand performs a circle with the radius of the object in
	the XZ-plane which yields a vertical
	rotation.
	\item \emph{Simple grasping}: The gripper is positioned on top of the object
	and the fingers are closed.
\end{itemize}
The haptic database consists of at least 10 samples per perceptual state.
Before sensing, the object is pushed back to a position in front of the robot.
We use four sensing actions:
\begin{itemize}
	\item \emph{No Sensing}: Some tasks do not require any prior sensing and have
	only one state.
	The discrimination score is computed with a success rate of $r_s = 0.5$,
	c.f. equation~\ref{equ:discscore}.
	\item \emph{Slide}: A finger is placed in front of the object.
	The object is pushed towards the finger with the second hand until contact
	or until the hands get too close to each other (safety reasons).
	Sensig is done by bending the finger.
	\item \emph{Press}: The object is pushed with one hand towards the second hand
	until the force exceeds a certain threshold.
	\item \emph{Poke}: The object is poked from the top with a finger.
	\item \emph{Weigh}: Checks a successful grasp by measuring the $z$-component of the Cartesian force..
	The perceptual states are fixed, i.e. not grasped~/~grasped.
\end{itemize}
\subsection{Real-world tasks}
\begin{figure*}[t!]
  \centering
  \subfloat[Without active learning and creativity]{\label{fig:comporiginal}\includegraphics[width=0.48\textwidth]{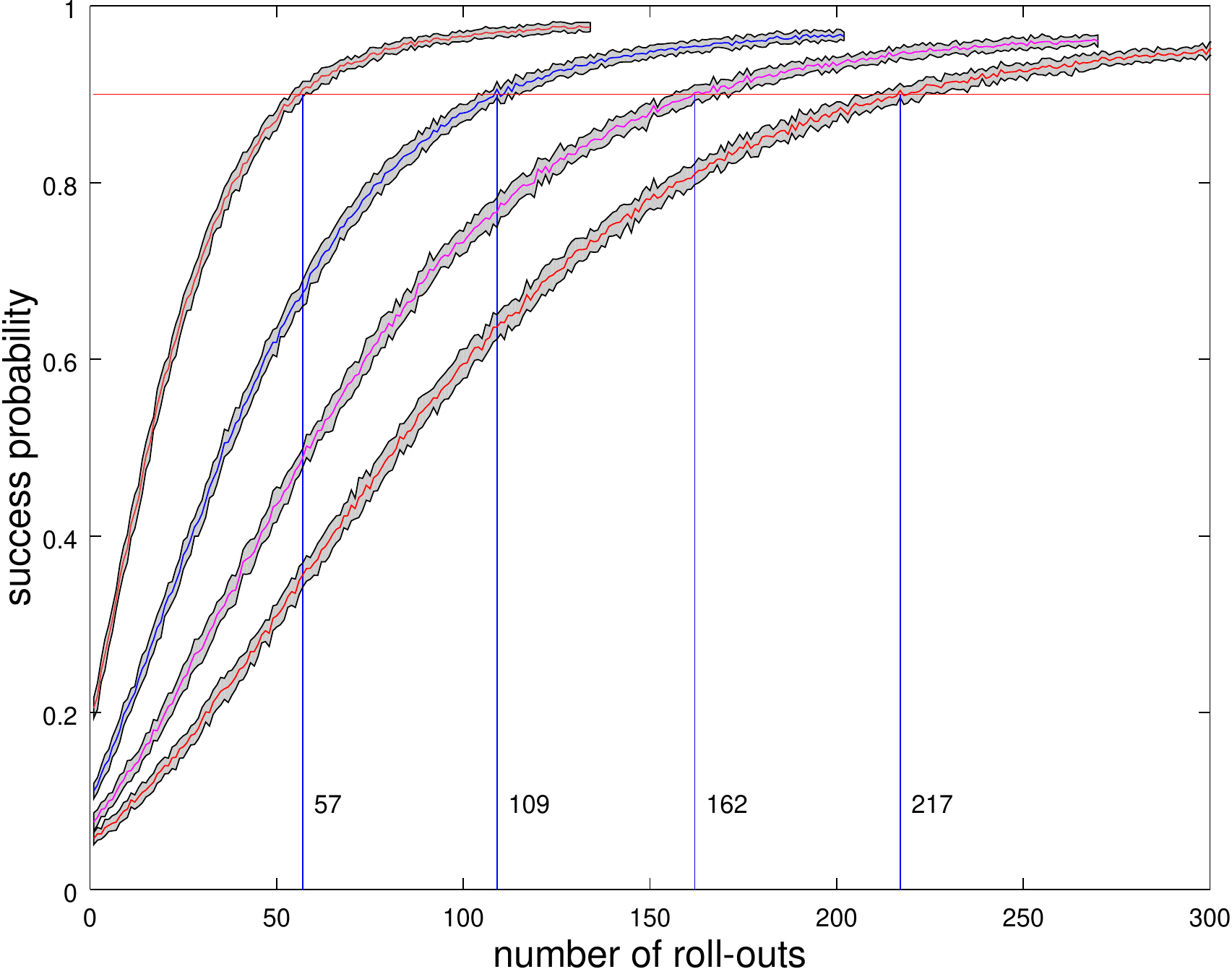}}
  \qquad
  \subfloat[With active learning and without creativity]{\label{fig:compboredom}\includegraphics[width=0.48\textwidth]{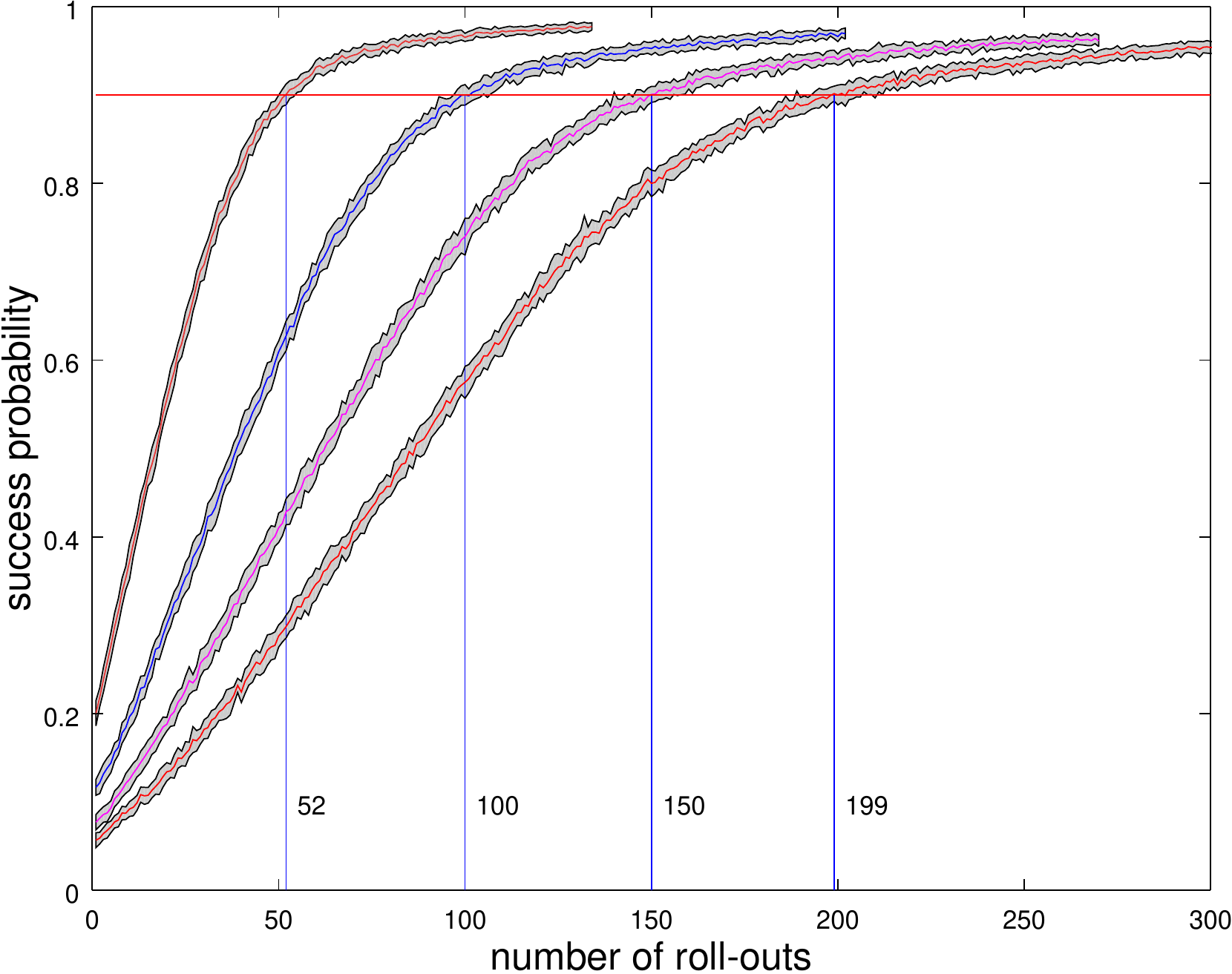}}
  \qquad
  \subfloat[With active learning and creativity]{\label{fig:compcreativity}\includegraphics[width=0.48\textwidth]{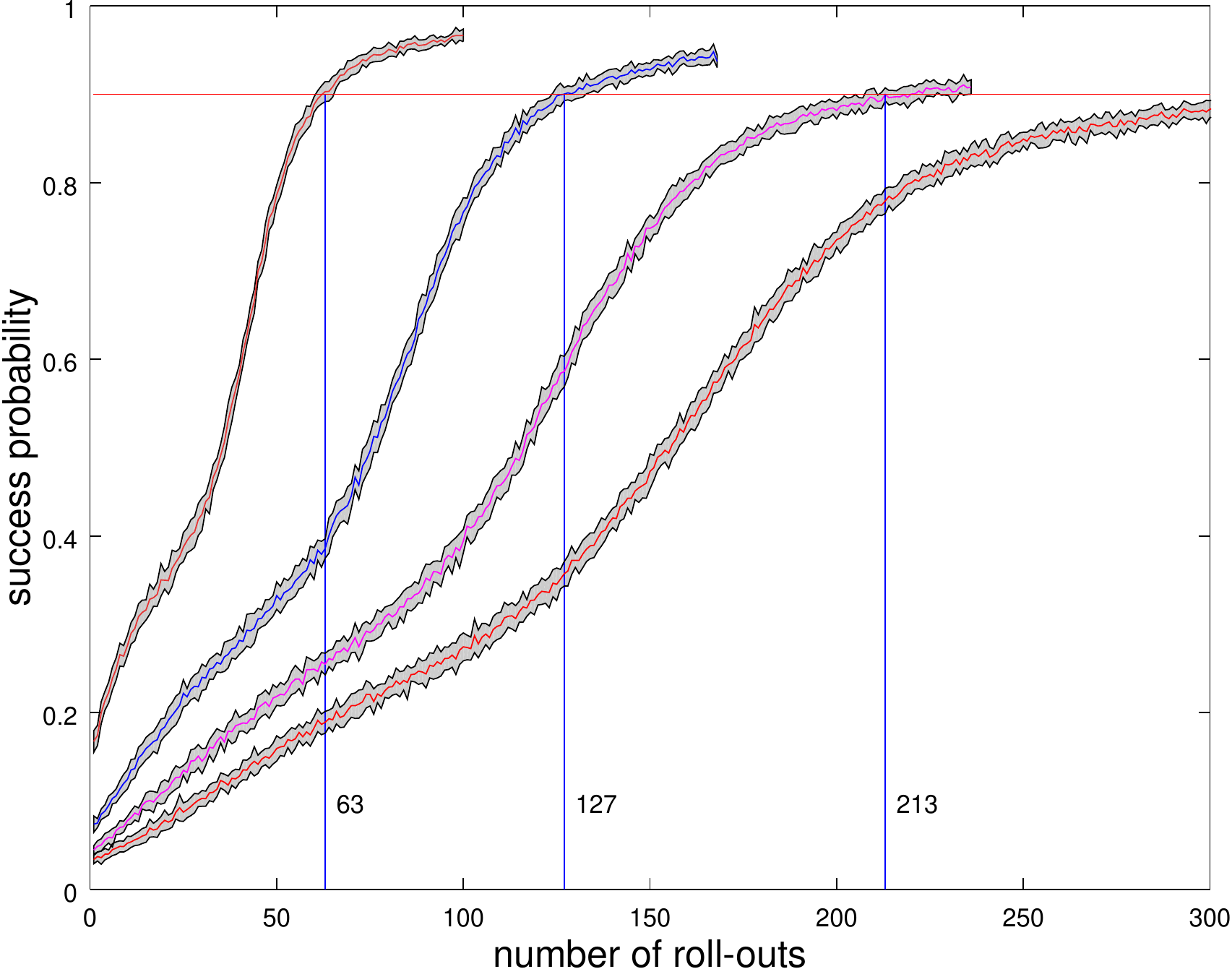}}
  \qquad
  \subfloat[Magenta $\equiv$ without active learning and creativity, green $\equiv$ with active learning and without creativity, cyan $\equiv$ with active learning and creativity]{\label{fig:compconvergence}\includegraphics[width=0.48\textwidth]{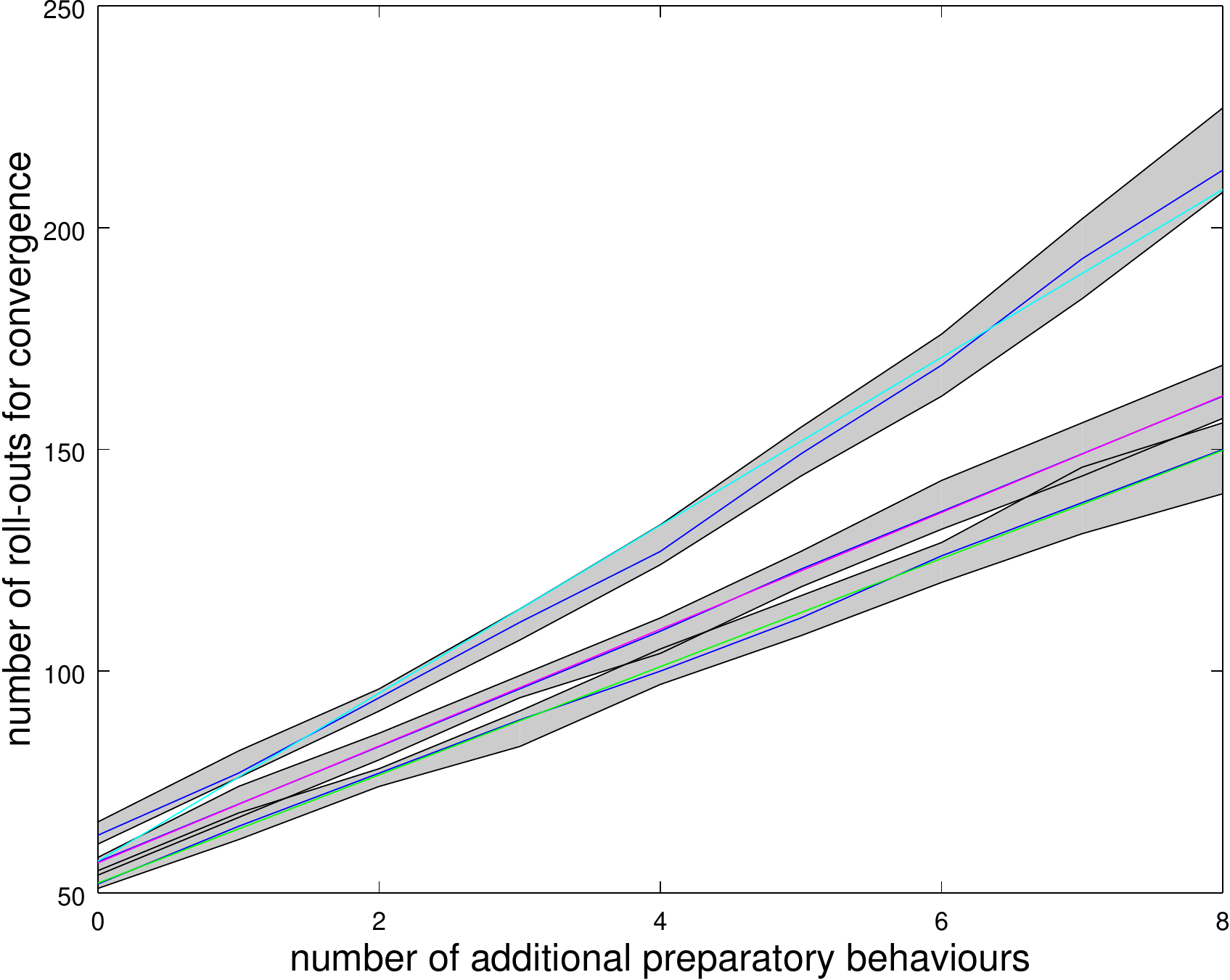}}
\caption{Figs. \ref{fig:comporiginal}~--~\ref{fig:compcreativity}: Evolution of the success rate over the number of rollouts for different numbers of preparatory behaviours.
For each curve from left to right, 5 behaviours are added.
The red horizontal lines denote an average success rate of 90 percent.
Fig. \ref{fig:compconvergence} shows the number of roll-outs required to reach an average success rate of at least 90 percent
for the three different versions.
The straight lines show a linear fit to the measured data.
}
\label{fig:compall}
\end{figure*}
We demonstrate the generality of our method in several scenarios.
Each skill can use the described preparatory behaviours, and
additionally, the skills trained before.
If not stated otherwise, all basic behaviours are
dynamic movement primitives (DMPs) \cite{schaal2006dynamic} trained by kinesthetic teaching.
A video of the trained skills including a visualisation of the generated skill hierarchies
can be viewed online\footnote{\url{https://iis.uibk.ac.at/public/shangl/tro2017/hangl\_roboticplaying.mp4}}
and is included in the supplementary material of this paper.
Note that only the skills and behaviours with non-zero probabilities are shown in the hierarchies.
The training of skills does not look different to the training in the basic method except
for the additional execution sensing action after the performed
preparation\footnote{\url{https://iis.uibk.ac.at/public/shangl/iros2016/iros.mpg}}.
\subsubsection{Simple placement}
The task is to pick an object and place it in an open box on the table.
The basic behaviour is a DMP that moves the grasped object to the box, where
the hand is opened.
In this case, the used sensing action is \emph{weigh}, c.f. \ref{sec:ecmplayingform}.
After training the \emph{simple grasp} / \emph{nothing} behaviour is used if the object is not grasped / not grasped respectively.
\subsubsection{Book grasping}
The basic behaviour grasps a book as described in section~\ref{sec:intro}.
The perceptual states are the four orientations of the book.
After training, the robot identified \emph{sliding} as a useful sensing action
to estimate the book's rotation.
The skill is trained with and without using creativity.
Without creativity, the available preparatory behaviours are the \emph{void}-behaviour,
\emph{rotate $90^{\circ}$}, \emph{rotate $180^{\circ}$}, \emph{rotate $270^{\circ}$},
and \emph{flip}.
The \emph{rotation} and \emph{void} behaviours are used
for different rotations of the book.
In the creativity condition, the behaviours \emph{rotate $180^{\circ}$}
and \emph{rotate $270^{\circ}$} are removed from the set
of preparatory behaviours.
The robot creates these behaviours by composing \emph{rotate $90^{\circ}$}
two / three times respectively.
\subsubsection{Placing object in a box}
The task is to place an object inside a box that can be closed.
The basic behaviour is to grasp an object from a fixed position
and drop it inside an open box.
The perceptual states determine, whether the box is
\emph{open} or \emph{closed}.
After training, the robot identifies \emph{poke} as a good sensing action.
The \emph{flip} behaviour is used to open the closed box and
the \emph{void}-behaviour is used if the box is open.
\subsubsection{Complex grasping}
The task is to grasp objects of different sizes.
We use the \emph{void}-behaviour as the skill's basic behaviour.
This causes the robot to combine behaviours without additional
input from the outside.
The perceptual states correspond to small and big objects.
After training, \emph{sliding} is determined as the best sensing action.
The \emph{simple grasp} / \emph{book grasping} behaviour is used for small / big
objects respectively.
\subsubsection{Shelf placement}
The task is to place an object in a shelving bay, which is executed
using a DMP.
The robot uses the \emph{weigh} sensing action to determine whether
or not an object is already grasped.
The \emph{complex grasp} skill / \emph{void} behaviour is used
if the object is not grasped / grasped, respectively.
Note that training of this skill can result in a
local maximum, e.g. by choosing the behaviours
\emph{simple grasp} or \emph{book grasp}, in particular if
the reward is chosen too high.
\subsubsection{Shelf alignment}
The task is to push an object on a shelf towards the wall to make space for more objects.
The basic behaviour is a DMP moving the hand from the left end of the shelve bay
to the right end until a certain force is exceeded.
As there is no object in front of the robot, all sensing
actions except \emph{no sensing} fail.
The sensing action with the strongest discrimination score
is \emph{no sensing} with only one perceptual state
and \emph{shelf placement} as preparatory behaviour.
\subsubsection{Tower disassembly}
The task is to disassemble a stack of maximum three boxes.
The basic behaviour is the \emph{void} behaviour.
The perceptual states correspond to number of boxes in the tower.
Reward is given in case the tower is completely disassembled.
After training, the used sensing action is \emph{poking} to
estimate four different states, i.e. height $h \in \{ 0, 1, 2, 3 \}$.
The tower cannot be removed with any single available preparatory behaviour.
Instead, using the creativity mechanism, the robot generates combinations of
\emph{simple placement}, \emph{shelf placement} and \emph{shelf alignment}
of the form given by the expression
\begin{equation}
\text{simple placement}^* \, [\text{void} \, | \, \text{shelf placement} \, | \, \text{shelf alignment}]
\label{equ:towerexperssions}
\end{equation}
\subsection{Discussion of the real-world tasks}
A strong advantage of model-free playing is the ability to use behaviours
beyond their initial purpose.
The \emph{flip} behaviour is implemented to flip an object but is used to
open the box in the \emph{box placement} task.
This holds for sensing actions as well:
\emph{sliding} is used for estimating the object
size for \emph{complex grasping} instead of the expected \emph{pressing}
from which the object size could be derived from the distance between the hands.
Both sensing actions deliver a high success rate with
$r_s^{\text{pressing}} \approx 0.9$ and $r_s^{\text{sliding}} \approx 1.0$.
The high success rate of \emph{sliding} is an artifact of the
measurement process.
The object is pushed towards the second hand
until the hands get too close to each other.
For small objects, the pushing stops before the finger
touches the object. 
This produces always the same sensor data for small objects,
which makes it easy to distinguish small from big objects.

In the \emph{tower disassembly} task an important propery can be observed.
The generated behaviour compositions of the form given in equation
\ref{equ:towerexperssions} only contain the skills \emph{shelf placement}
and \emph{shelf alignment} at the end of the sequence.
The reason is that these skills can only remove a box in a controlled way
if only one box is left, i.e. $h = 1$.
Higher towers are made to collaps because of the \emph{complex grasping} skill,
which is used by \emph{shelf placement}.
It uses \emph{sliding} to estimate the object's size and therefore pushes the
tower around.
Further, which behaviour sequence is generated, depends on the subjective
history of the robot, e.g. the sequences (\emph{simple placement}, \emph{simple placement},
\emph{simple placement}) and (\emph{shelf placement}, \emph{simple placement}, \emph{simple placement}) both yield success for $h = 3$.
The autonomy of our approach can also be reduced in such a scenario, as several
behaviours destroy the tower and require a human to prepare it again.
This involves to include a human in the playing loop, in particular if the required
states cannot be prepared by the robot itself.

Similarly, the active learning and creativity mechanisms do not
always yield improvements.
Active learning only causes a speedup if the
unsolved perceptual states can be produced from solved ones,
e.g. if the \emph{closed} state is solved before the \emph{open} state.
The robot is only able to prepare the transition
\emph{closed} $\xrightarrow{flip}$ \emph{open}.
The transition \emph{open} $\xrightarrow{}$ \emph{closed}
requires to close the cover, which is not among the
available behaviours.
The creativity mechanism does not improve learning
if the required behaviours are already available,
e.g. in \emph{box placement} or \emph{shelf placement},
or cannot be composed of other behaviours.
However, it helps to solve \emph{book grasping} and \emph{tower disassembly}.

We emphasise that the teaching of novel skills does not
necessarily have to follow the typical sequence of
\emph{sensing} $\rightarrow$ \emph{preparation} $\rightarrow$ \emph{basic behaviour},
e.g. in \emph{complex grasping} and \emph{shelf alignment}.
In the \emph{complex grasping} task the basic behaviour is the \emph{void}-behaviour,
which causes the robot to coordinate different grasping procedures for small and big objects.
For \emph{shelf alignment}, the sensing stage is ommitted.
\subsection{Simulated skill learning}
Single experiments cannot be used to assess the overall convergence behaviour.
We use the experiences gained in the real-world \emph{book grasping} task to simulate
the convergence behaviour.
We use a success rate of 95 percent for all involved controllers.
The environment is simulated with ground truth state transitions observed in the real-world experiment.
For the failure cases, i.e. 5 percent of the exeuctions of each executed action,
we simulate a random resulting perceptual state.
The evolution of success is simulated and averaged for $N = 1000$ robots for different
numbers of preparatory behaviours.
The minimum number of preparatory behaviours is $J = 5$, i.e. \emph{void},
\emph{rotate $90^{\circ}$}, \emph{rotate $180^{\circ}$}, \emph{rotate $270^{\circ}$}, \emph{flip}.
We simulate a scenario in which additional preparatory behaviours are useless,
i.e. the perceptual state is not changed.
In this case, the problem gets harder due to a larger set of behaviours, while
the number of appropriate behaviours remains the same.
In the scenario with activated creativity the agent is only provided with the 
behaviours \emph{rotate $90^{\circ}$}, \emph{flip} and \emph{void}.

The number of rollouts required to reach a success rate of at least 90 percent
is given in Table \ref{tab:convs} for an increasing number of behaviours $J$ and
different variants of our method
($N_{\text{no\_ext}} \equiv$ no active learning / no creativity,
$N_{\text{active}} \equiv$ active learning / no creativity,
$N_{\text{creative}} \equiv$ active learning / creativity,
$N_{\text{base}} \equiv$ baseline).
As baseline we use a policy in which every
combination of perceptual states and behaviours is tried out only once,
with $N_{\text{base}} = 3 * 4 * J + J$ (3 sensing actions with 4 states,
1 sensing action, i.e. \emph{no sensing} with only one state).
In general, our method converges faster than the baseline due to reducing
the space strongly and ignoring irrelevant parts of the ECM.
Further, the baseline method would not yield convergence in a scenario with
possible execution failures as each combination is executed only once.
The baseline approach also cannot solve the
task in the creativity condition.

\begin{table}[t!]
\caption{
Number of rollouts required to converge to a success rate of at least 90 percent
for different numbers of behaviours $J$.
}
\center
\begin{tabular}{lllll}
\hline
$J$ & $N_{\text{no\_ext}}$ & $N_{\text{active}}$ & $N_{\text{creative}}$ & $N_{\text{base}}$ \\
\hline
5 & 57 & 52 & 63 & 65 \\
10 & 109 & 100 & 127 & 130 \\
15 & 162 & 150 & 213 & 190 \\
20 & 217 & 199 & $>$ 300 & 260 \\
\hline
\end{tabular}
\label{tab:convs}
\end{table}

The two versions without creativity, i.e. without and with active learning,
show continuous increase of the success rate in
Figs. \ref{fig:comporiginal} and \ref{fig:compboredom}.
If the robot is bored, situations with a low information gain are rejected.
Therefore, the version with active learning is expected to converge faster.
Fig. \ref{fig:compconvergence} shows the number of required rollouts to
reach a success rate of 90 percent for each of the three variants.
The number of required rollouts is proportional to the number of available preparatory behaviours.
We apply a linear fit and gain an asymptotic speed-up of
$sp = 1 - \lim_{x \rightarrow \infty} \frac{k_1 x + d_1}{k_2 x + d_2}
\approx	9$ percent for the variant with active learning compared
to the variant without extension.

In the scenario with activated creativity the convergence behaviour
is different, c.f. Fig. \ref{fig:compcreativity}.
The success rate exhibits a slow start followed by a fast increase
and a slow convergence towards 100 percent.
The slow start is due to the perceptual
states that would require the behaviours \emph{rotate $180^{\circ}$} and \emph{rotate $270^{\circ}$}
which are not avaible at this point.
Further, the robot cannot generate these behaviours using
creativity due to initially untrained environment models.
This causes the success rate to reach a preliminary plateau at around
30 to 35 percent.
After this initial burn-in phase, the environment model becomes
more mature and behaviour proposals are created.
This causes a strong increase of the the success rate.
\section{Conclusion}
We introduced a novel way of combining model-free and
model-based reinforcement learning methods for autonomous skill acquisition.
Our method acquires novel skills that work for only a
narrow range of situations acquired from a human teacher, e.g. by demonstration.
Previously-trained behaviours are used in a model-free RL setting in order
to prepare these situations from other possibly occuring ones.
This enables the robot to extend the domain of applicability of the novel skill
by playing with the object.
We extended the model-free approach by learning an environment model
as a side product of playing.
We demonstrated that the environment model can be used to improve the model-free
playing in two scenarios, i.e. active learning and creative behaviour generation.
In the active learning setting the robot has the choice of rejecting present situations
if they are already well-known.
It uses the environment model to autonomously prepare more interesting situations.
Further, the environment model can be used to propose novel preparatory behaviours
by concatenation of known behaviours.
This allows the agent to try out complex behaviour sequences while still preserving
the model-free nature of the original approach.

We evaluated our approach on a KUKA robot by solving complex manipulation tasks, e.g.
complex pick-and-place operations, involving non-trivial manipulation, or tower-disassembly.
We observed success statistics of the involved components and simulated the convergence behaviour
in increasingly complex domains, i.e. a growing number of preparatory behaviours.
We found that by active learning the number of required
rollouts can be reduced by approximately 9 percent.
We have shown that creative behaviour generation enables the robot to solve
tasks that would not have been solvable otherwise,
e.g. complex book grasping with a reduced number of preparatory
behaviours or tower disassembly.

The work presented in this paper bridges the gap from plain
concatenation of pre-trained behaviours to simple goal-directed planning.
This can be seen as early developmental stages of a robot.
We believe that a lifelong learning agent has to go through different stages
of development with an increasing complexity of knowledge and improving
reasoning abilities.
This raises the question of how the transition to strong high-level
planning systems could look like.

Our experiments show that the learning time is proportional to
the number of used preparatory behaviours.
This makes it efficient to learn an initial (and potentially strong)
set of skills, but hard to add more skills when there is a large set
of skills available already.
Training more sophisticated models could help to overcome this problem.
Further, in the current system, the creative behaviour generation
only allows behaviour compositions resulting from plans within the same
environment model, i.e. using only perceptual states of the same sensing action.
The expressive power of our method could be greatly increased by allowing
plans through perceptual states of different sensing actions.
This could also involve multiple sensing actions at the same time including passive
sensing such as vision.
\section*{Acknowledgment}
The research leading to these results has received funding
from the European Communitys Seventh Framework Programme
FP7/20072013 (Specific Programme Cooperation, Theme 3,
Information and Communication Technologies) under grant
agreement no. 610532, Squirrel. HJB and VD acknowledge
support from the Austrian Science Fund (FWF) through
grant SFB FoQuS F4012.

\ifCLASSOPTIONcaptionsoff
  \newpage
\fi

\bibliographystyle{IEEEtran}
\bibliography{IEEEabrv,bibfile}

\begin{IEEEbiography}[{\includegraphics[width=1in,height=1.25in,clip,keepaspectratio]{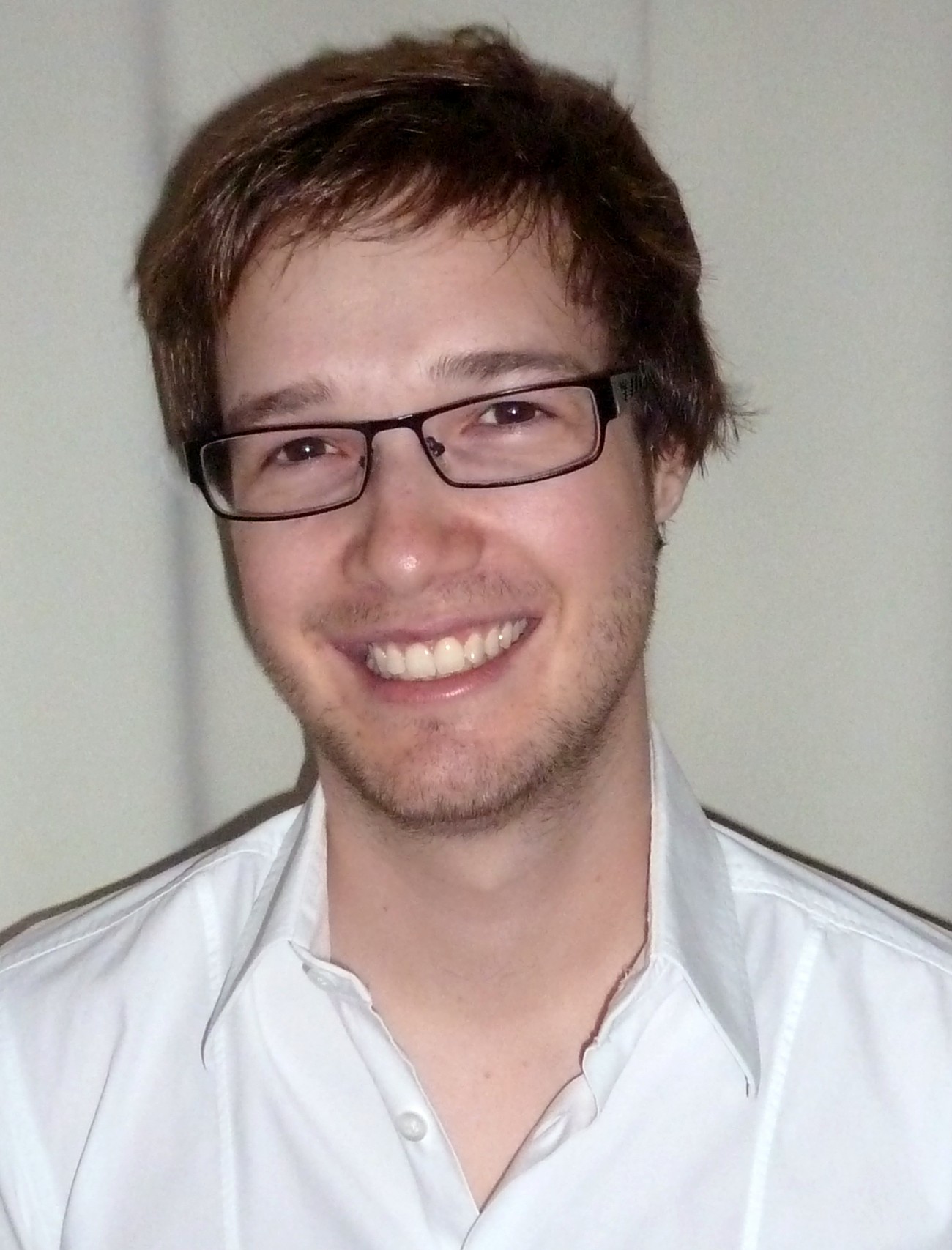}}]{Simon Hangl}
is a PhD student at the intelligent and interactive systems (IIS) group of the University of Innsbruck, where he also received his MSc degree in Computer
Science. Before starting his position as University Assistant at IIS, he worked as researcher at the Semantic Technology Institute (STI) Innsbruck. He worked in
4 EU-FP7 projects in the areas of semantic technology and robotics. He is interested in developmental and cognitive robotics and complex robotic object manipulation.
\end{IEEEbiography}
\begin{IEEEbiography}[{\includegraphics[width=1in,height=1.25in,clip,keepaspectratio]{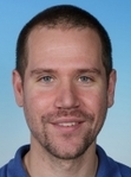}}]{Vedran Dunjko} received the Ph.D. degree in
physics from Heriot-Watt University, Edinburgh,
U.K., in 2012, focused on problems in quantum
cryptography, specifically blind quantum computing,
and quantum digital signatures. Following a
one-year post-doctoral position with the School of
Informatics, University of Edinburgh, in 2013, he
then moved to the Institute of Theoretical Physics,
University of Innsbruck, Austria, where he is currently
a Post-Doctoral Researcher. He has recently
been involved in the problems of artificial intelligence and quantum machine
learning.
\end{IEEEbiography}
\begin{IEEEbiography}[{\includegraphics[width=1in,height=1.25in,clip,keepaspectratio]{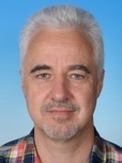}}]{Hans J. Briegel} received the Ph.D. degree
(doctorate) in physics in 1994 and the Habilitation
degree in theoretical physics in 2002 from
the Ludwig-Maximilians-University of Munich.
He held postdoctoral positions with Texas A\&M
and Harvard University. He has been a Full Professor
of Theoretical Physics with the University
of Innsbruck since 2003 and a Research Director
with the Institute of Quantum Optics and Quantum
Information of the Austrian Academy of Sciences
from 2003 to 2014. His main field of research is quantum information and
quantum optics where he has authored and co-authored papers on a wide
range of topics, including work on noise reduction and microscopic lasers,
quantum repeaters for long-distance quantum communication, entanglement
and cluster states, and measurement-based quantum computers. His recent
research has focussed on physical models for classical and quantum machine
learning, artificial intelligence, and the problem of agency.
HJ Briegel is also with the Department of Philosophy, University of Konstanz, Germany.
\end{IEEEbiography}
\begin{IEEEbiography}[{\includegraphics[width=1in,height=1.25in,clip,keepaspectratio]{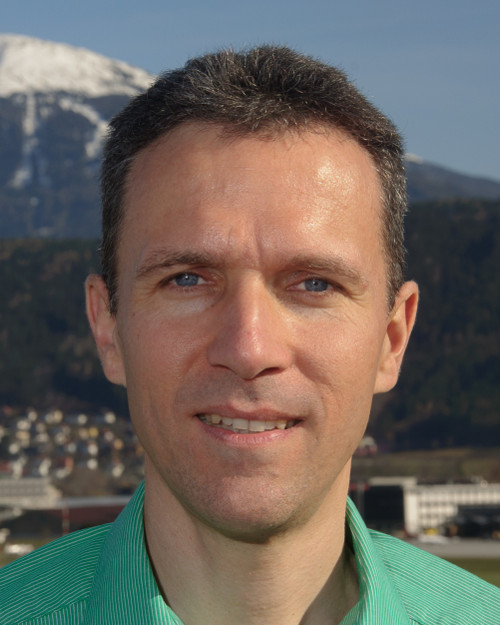}}]{Justus Piater} is a professor of computer science at the University of
Innsbruck, Austria, where he leads the Intelligent and Interactive
Systems group. He holds a M.Sc. degree from the University of
Magdeburg, Germany, and M.Sc. and Ph.D. degrees from the University of
Massachusetts Amherst, USA, all in computer science.  Before joining
the University of Innsbruck in 2010, he was a visiting researcher at
the Max Planck Institute for Biological Cybernetics in T\"ubingen,
Germany, a professor of computer science at the University of Liège,
Belgium, and a Marie-Curie research fellow at GRAVIR-IMAG, INRIA
Rhône-Alpes, France.  His research interests focus on visual
perception, learning and inference in sensorimotor systems.  He has
published more than 150 papers in international journals and
conferences, several of which have received best-paper awards, and
currently serves as Associate Editor of the IEEE Transactions on
Robotics.
\end{IEEEbiography}

\vfill

\end{document}